\documentclass{article}

\usepackage{microtype}
\usepackage{graphicx}
\usepackage{subcaption}
\usepackage{booktabs} 

\usepackage{hyperref}
\usepackage{algorithm}

\usepackage[preprint]{icml2026}

\usepackage{amsmath}
\usepackage{amssymb}
\usepackage{mathtools}
\usepackage{amsthm}
\usepackage{makecell}
\usepackage{placeins}
\usepackage[capitalize,noabbrev]{cleveref}
\usepackage[table]{xcolor}

\usepackage{multirow}
\usepackage{wrapfig}

\theoremstyle{plain}

\theoremstyle{definition}

\theoremstyle{remark}

\usepackage[textsize=tiny]{todonotes}
\usepackage{float}
\usepackage{url}

\definecolor{Detection}{RGB}{150,200,250}
\definecolor{ComplexTactic}{RGB}{200,150,250}
\definecolor{SourceTargetSquare}{RGB}{250,200,150}
\definecolor{Value}{RGB}{170,180,120} 
\definecolor{Capture}{RGB}{210,140,140} 
\definecolor{PieceMoving}{RGB}{130,190,180}
\definecolor{Space}{RGB}{170,190,245} 

\definecolor{DetectionBG}{RGB}{235,245,255}
\definecolor{ComplexTacticBG}{RGB}{245,235,255}
\definecolor{SourceTargetSquareBG}{RGB}{255,245,230}
\definecolor{ValueBG}{RGB}{242,244,232}
\definecolor{CaptureBG}{RGB}{248,236,236}
\definecolor{PieceMovingBG}{RGB}{232,246,244}
\definecolor{SpaceBG}{RGB}{240,242,252}

\definecolor{BestMove}{HTML}{2F9E44}
\definecolor{Suboptimal}{HTML}{E03131}
\definecolor{Alternative}{HTML}{9C36B5}

\usepackage{titlesec}
\titlespacing*{\paragraph}{0pt}{0.5ex plus .2ex minus .1ex}{1em}

\begin{document}

\twocolumn[
    \icmltitle{Tracing the Thought of a Grandmaster-level Chess-Playing Transformer}


  \icmlsetsymbol{equal}{*}

  \begin{icmlauthorlist}
    \icmlauthor{Rui Lin}{yyy,sch}
    \icmlauthor{Zhenyu Jin}{comp}
    \icmlauthor{Guancheng Zhou}{yyy,comp}
    \icmlauthor{Xuyang Ge}{yyy,sch}
    \icmlauthor{Wentao Shu}{yyy,sch}
    \icmlauthor{Jiaxing Wu}{yyy,sch}
    \icmlauthor{Junxuan Wang}{yyy,sch}
    \icmlauthor{Zhengfu He}{yyy,sch}
    \icmlauthor{Junping Zhang}{sch}
    \icmlauthor{Xipeng Qiu}{yyy,sch}
  \end{icmlauthorlist}

  \icmlaffiliation{yyy}{Shanghai Innovation Institute, Shanghai, China}
  \icmlaffiliation{sch}{School of Computer Science, Fudan University, Shanghai, China}
  \icmlaffiliation{comp}{School of Mathematics and Statistics, Xi’an Jiaotong University}
  
  \icmlcorrespondingauthor{Xipeng Qiu}{xpqiu@fudan.edu.cn}

  \icmlkeywords{Mechanistic Interpretability, SAE, Chess}

  \vskip 0.3in
]



\printAffiliationsAndNotice{}  

\begin{abstract}
  While modern transformer neural networks achieve grandmaster-level performance in chess and other reasoning tasks, their internal computation process remains largely opaque. 
  Focusing on Leela Chess Zero (LC0), we introduce a sparse decomposition framework to interpret its internal computation by decomposing its MLP and attention modules with sparse replacement layers, which capture the primary computation process of LC0.
  We conduct a detailed case study showing that these pathways expose rich, interpretable tactical considerations that are empirically verifiable.
  We further introduce three quantitative metrics and show that LC0 exhibits parallel reasoning behavior consistent with the inductive bias of its policy head architecture.
  To the best of our knowledge, this is the first work to decompose the internal computation of a transformer on both MLP and attention modules for interpretability.
  Combining sparse replacement layers and causal interventions in LC0 provides a comprehensive understanding of advanced tactical reasoning, offering critical insights into the underlying mechanisms of superhuman systems. Our code is available at \url{https://github.com/JacklE0niden/Leela-SAEs}.
\end{abstract}

\section{Introduction}

\begin{figure}[t]
    \centering
    \includegraphics[width=0.99\linewidth]{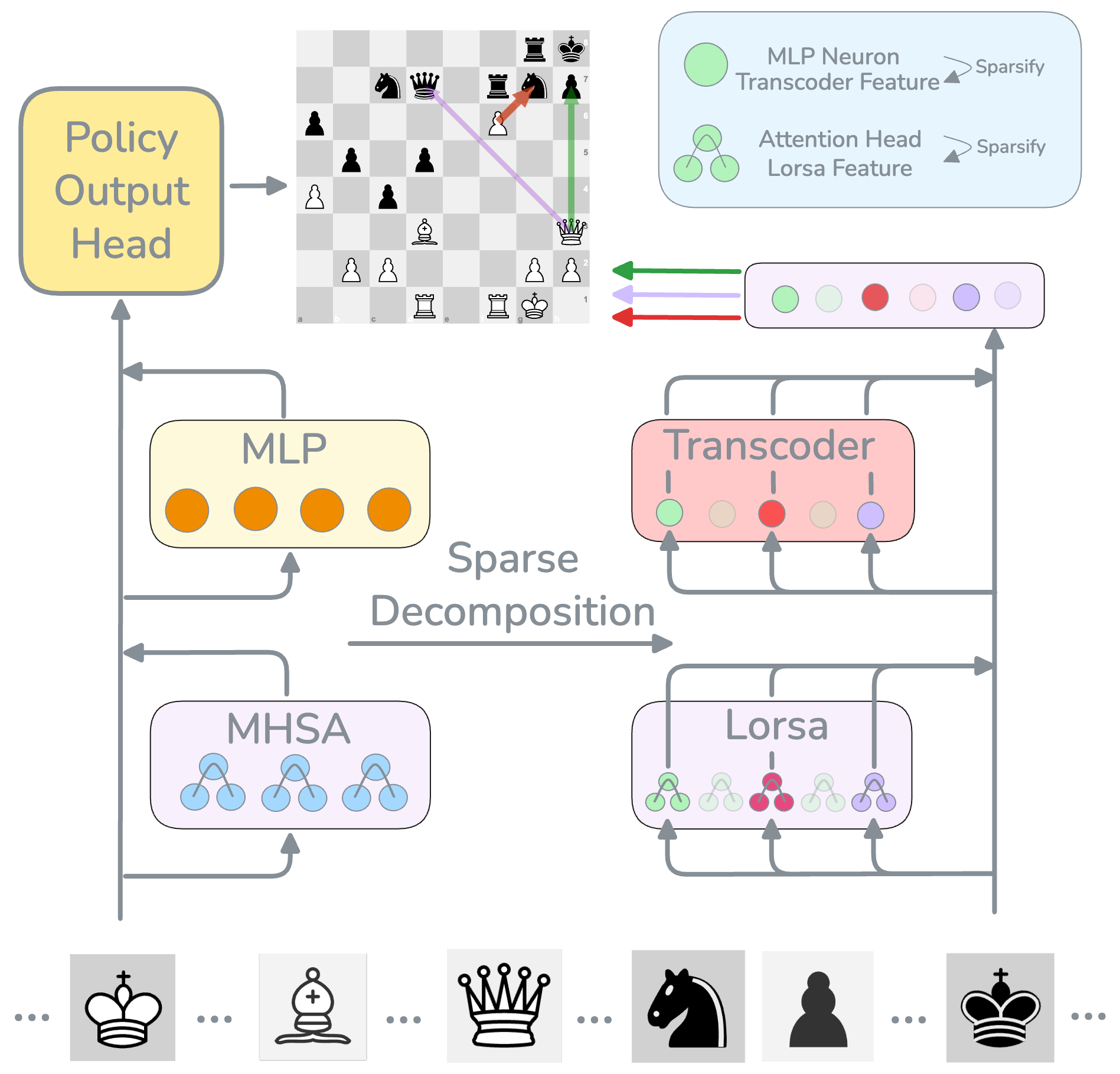}
    \caption{
    An overview of our approach for interpreting LC0. We use Transcoders and Lorsas as replacement layers to sparsely decompose the model's MLP and attention module, identifying interpretable reasoning pathways through feature-steering interventions. By analyzing sparse feature activations along these pathways, we reveal an interpretable computational process underlying the model's decision-making for any input position.
    }
    \label{fig:head}
\end{figure}

Modern neural networks have demonstrated remarkable abilities in reasoning tasks such as board games~\cite{tesauro1994td, campbell2002deep, silver2016mastering, silver2017mastering}, mathematics~\cite{seo2015solving, lightman2023let, 
romera2024mathematical}, and coding~\cite{li2022competition}. 
Building on the MCTS-based self-play training framework~\cite{silver2017mastering}, modern Transformer-based models have achieved remarkable playing strength in chess.
Among them, the \textbf{Leela Chess Zero (LC0)} series achieves grandmaster-level performance even \textbf{without search}~\cite{ruoss2024amortized, lczero2024transformer, monro2024mastering}.

Despite their success, the understanding of LC0's internal computation is still incomplete: prior studies have focused on local concepts~\cite{mcgrath2022acquisition} or specific attention heads~\cite{jenner2024evidence}, while a comprehensive, global technique to track computation across the entire model is still missing. 
A thorough interpretability study is thus crucial both to gain principled insights into the advanced reasoning of superhuman systems and to extract knowledge from their internal computations.

While model representations and computations are entangled in superposition and attention superposition~\cite{elhage2022tms, jermyn2024attentionsuperposition, lindsey2024crosscoder}, individual neurons are uninterpretable. Replacement layers based on sparse decomposition like Transcoders~\cite{ge2024hierattr, dunefsky2024transcoder, ameisen2025circuit} have proven effective for extracting features from the computational superposition~\cite{dunefsky2024transcoder, he2025towards}.

We utilized Transcoders and Low-Rank Sparse Attention modules (Lorsa)~\cite{he2025towards} to decompose MLPs and attentions into interpretable, feature-level computational units (which we call features). By integrating them into a causal intervention framework via feature steering~\cite{bricken2023monosemanticity, chalnev2024improving}, we introduce \textbf{reasoning pathways} and a method for constructing them to capture the internal computation process of the model. To our knowledge, this represents the first attempt to decompose both MLP and attention modules of a transformer model using unsupervised sparse dictionary learning techniques~\cite{kreutz2003dictionary, bricken2023monosemanticity}, and thus to enable full-model circuit-level analysis with these sparse replacement layers.

Through a detailed case study on LC0, we show that reasoning pathways reveal internal tactical reasoning processes, such as transferring offensive information and the opponent's defensive coverage across squares. In addition, our analysis diagnoses that the ambiguity in that position can be traced to
an over-evaluation of defensive dependencies. These mechanistic findings are supported by targeted interventions. Overall, features in and reasoning pathways facilitate a mechanistic understanding of how LC0 derives tactical reasoning logic from the input position.

To characterize reasoning pathways, we introduce three metrics — \textbf{path overlap} ($\bar{O}$), \textbf{path cohesion} ($\mathcal{C}_{oh}$) and \textbf{path coupling} ($\mathcal{C}_{up}$) to quantify the parallelism of pathways for different moves. We further show, across layers, critical representations progressively converge to source and target squares corresponding to the move, aligning with LC0’s policy head architecture.

By leveraging reasoning pathways constructed with sparse replacement layers, our approach offers a comprehensive technique to trace the internal reasoning across an entire transformer. Our findings in LC0 also provide critical insight into superhuman reasoning systems and how humans might extract knowledge from inside~\cite{shin2023superhuman}.

\paragraph{Roadmap.}
In Section~\ref{sec:preliminaries}, we introduce LC0, Transcoder, and Lorsa, and describe a visualization interface for inspecting features as well as rule-based verification applied to a subset of interpretable features.
In Section~\ref{sec:discovering_reasoning_pathways}, we introduce our method to construct reasoning pathways for any input chess position and a targeted move.
In Section~\ref{sec:case_study}, we conduct an in-depth case study to decompose LC0's reasoning pathways, revealing computational features in pathways that execute tactic logics such as offensive planning and defensive evaluation. We further diagnose a pathological reasoning state related to over-evaluation of defensive constraints.
In Section~\ref{sec:dive_into_thinking_paths}, we introduce quantitative metrics to verify that reasoning pathways for different candidate moves are highly parallel and distributed, and the model forms a progressively aggregated representation for a move.

Our main contributions are as follows.
\begin{itemize}
    \item We adapt Transcoders and Lorsas to LC0 BT4 and, to our knowledge, present the first sparse decomposition of a full transformer, extracting interpretable computational features and providing the first \textbf{rule-based validation} of their alignment with theoretical chess concepts in LC0.
    
    \item We propose a feature-steering-based method for constructing \textbf{reasoning pathways} in LC0, demonstrating through case studies that these pathways reveal the rich tactical logic within the model’s internal computations.
    
    \item We introduce three quantitative metrics to characterize the \textbf{parallelism} of LC0's reasoning pathways, and show that decision-critical information \textbf{progressively converges} onto a move's source and target squares.
\end{itemize}

\section{Preliminaries}
\label{sec:preliminaries}

\subsection{Sparse Decomposition via Replacement Layers}
To disentangle computations under superposition~\cite{elhage2022tms, jermyn2024attentionsuperposition}, we employ a \emph{replacement layer} approach~\cite{ameisen2025circuit} by applying Transcoders for MLPs and Lorsas~\cite{he2025towards} for attention modules. This allows for the extraction of monosemantic features within the model’s primary computational components.

\paragraph{Transcoder.}
A Transcoder approximates the computation of an MLP layer using a sparse, linear decomposition.
Given the MLP input $x \in \mathbb{R}^d$, the Transcoder encoder produces sparse feature activations via a Top-$K$ operation:
\begin{equation}
z = \mathrm{Top}\text{-}K(W_E x + b_E),
\end{equation}
where $z \in \mathbb{R}^m$ contains only the $K$ largest activations, with all other entries set to zero.
The MLP output is then reconstructed as a linear combination of decoder vectors:
\begin{equation}
\hat{y} = W_D z + b_D.
\end{equation}
This formulation mimics the MLP's behavior while allowing for individual feature analysis and causal intervention.

\paragraph{Low-rank Sparse Attention (Lorsa).}
A Lorsa disentangles attention heads in superposition into linear rank-1 \emph{output value (OV)} heads, which can be interpreted as a sparse dictionary of features. 
Sparsity constraints ensure that only a few heads are activated per token position. 
We fix the number of Lorsa heads to match the number of Transcoder features, and for consistency, we denote the ``Lorsa heads'' in~\cite{he2025towards} as ``Lorsa features''.

Formally, each Lorsa feature takes in $x \in \mathbb{R}^{n \times d}$ and computes an attention following multi-head self-attention (MHSA), but with a 1-d value, i.e.,
\begin{equation}
\begin{aligned}
A^h &= \mathrm{softmax}\Bigg(
\frac{x W^h_Q (x W^h_K)^\top}{\sqrt{d_h}}
\Bigg) \in \mathbb{R}^{n \times n}, \\
v^h &= x w^h_V \in \mathbb{R}^{n \times 1} \; \text{(scalar value per token)}
\end{aligned}
\end{equation}

where $W^h_Q, W^h_K \in \mathbb{R}^{d \times d_h}$ and $w^h_V \in \mathbb{R}^{d \times 1}$ are weights.

The scalar activation of a Lorsa feature at token position $i$ is computed as
\begin{equation}
z_i^h = A^h_i v^h = \sum_{j=1}^{n} A^h_{i,j} \,(w^h_V)^\top x_j \in \mathbb{R}^{n \times 1},
\label{eq:compute_z}
\end{equation}
which can be decomposed into token-wise contributions, or \emph{$z$-patterns}, 
\begin{equation*}
z\text{-pattern}_{i,j} = A^h_{i,j} (w^h_V)^\top x_j,
\end{equation*}
indicating how strongly token $j$ contributes to the activation at position $i$. 
The feature output is then given by $\hat{Y}_h = z^h W^h_O$, with $W^h_O$ playing a role analogous to the decoder weight $W_D$ in Transcoders.

\paragraph{Dataset and Training.} 
We use chess positions from the \texttt{lichess\_standard\_rated}\footnotemark dataset to generate activations and train both Transcoders and Lorsas.

While each position is represented as a sequence of 64 tokens corresponding to 64 squares, we sample 800M tokens for training and 100M for analysis of the Transcoders and Lorsas. Training and adaptation details are provided in Appendix~\ref{appendix:sparse_decomposition_details}.

\footnotetext{The \texttt{lichess\_standard\_rated} dataset is publicly available at \url{https://database.lichess.org/} and contains over 7.4 billion rated standard chess games played on \texttt{lichess.org}.}

\subsection{Measuring Interpretability for Features}
\label{sec:interpretability}

\begin{figure}
    \centering
    \includegraphics[width=0.99\linewidth]{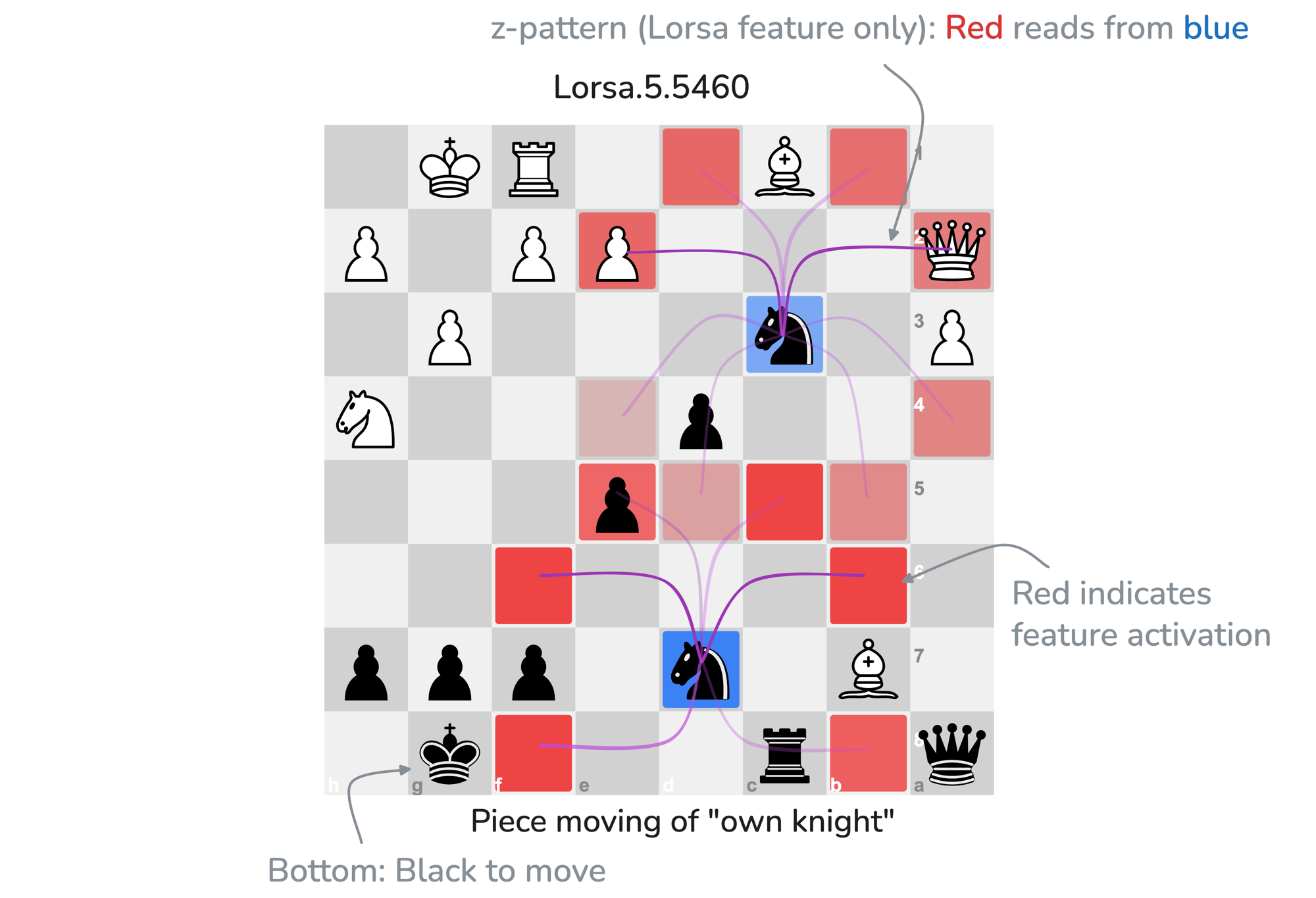}
    \caption{\textbf{Visualization interface for feature interpretability.} Red squares indicate the spatial activations of specific features. For Lorsa features, each activated square is associated with a $z$-pattern, highlighting its attentional focus (e.g., the black knight, marked in blue). Purple arrows connect each activated square to the target square to which its activation is highly attributed.}
    \label{fig:interface}
\end{figure}

\paragraph{Interpretability Interface.} 
We interpret Transcoder and Lorsa features via their highest-activating board squares and validate them using \textbf{rule-based statistical tests}, leveraging the structured nature of chess. Table~\ref{tab:feature_validation} presents a categorized, non-exhaustive feature set along with precision and recall of the verifications. For visualization, Transcoder features show scalar activations at each square, while Lorsa features display $z$-patterns~\cite{he2025towards}. Figure~\ref{fig:interface} provides a unified view of feature activations, including source squares for Lorsa features.

\paragraph{Notation.} We denote a feature instance as \textbf{\texttt{Tc.$\ell$.f@$s$}} for Transcoder features and \textbf{\texttt{Lorsa.$\ell$.f@$s$}} for Lorsa features. For example, \textbf{\texttt{Lorsa.7.14502@h7}} refers to Lorsa feature \#14502 of layer~7, whose activation is on \textbf{\texttt{h7}} square. We give a necessary explanation in Appendix~\ref{appendix:interpretation_explain} to clarify our interpretation language for features.

\begin{table}[H]
\centering
\caption{
    Interpretability assessment. Human evaluation scores (1--5 scale) by three player annotators (Elo 2100+, 2000+, and 1000+), measuring activation consistency and perceived complexity across different feature types. 
    %
}
\label{tab:feature_human_eval}
\small
\begin{tabular}{lcc}
\toprule
\textbf{Feature Type} & \textbf{Activation Consistency} & \textbf{Complexity} \\
\midrule
Transcoder & \textbf{4.15} & 2.25 \\
Lorsa      & 3.73 & \textbf{2.54} \\
MLP        & 2.05 & 1.45 \\
\bottomrule
\end{tabular}

\end{table}

\begin{table*}[t]
  \centering
  \caption{
    We present a categorized collection of interpretable Transcoder and Lorsa features, organized by coarse semantic function rather than architectural layer.
    Features categories are color-coded as follows:
    {\color{Detection}Detection Features([Det])},
    {\color{SourceTargetSquare}Source/Target Square Features([Src]/[Tgt])},
    {\color{Value}Value Features([Val])},
    {\color{Capture}Capture Features([Cap])},
    {\color{ComplexTactic}Complex Tactic Features([Tac])},
    {\color{Space}Spatial Relation Features([Spa])},
    {\color{PieceMoving}Piece Movement Features([Mov])}.
    Each feature is validated using a manually designed rule, and precision and recall are computed over training dataset.
    Detailed explanations of the interpretation language and verification rules are provided in Appendix~\ref{appendix:interpretation_explain}.
    We cap the analysis of the 1000 activation times per feature in a random dataset and compute precision and recall based on these samples.
    Detailed explanations of the interpretation language and verification rules are provided in Appendix~\ref{appendix:interpretation_explain}, where we present an illustrative visualization with a single representative feature per category to convey the core semantic pattern in Appendix~\ref{appendix:taxonomy_feature}.
    \textbf{Bolded features} indicate those used in the following case study. While some features exhibit lower recall due to feature splitting~\cite{chanin2025absorption}—where descriptive rules are coarser than actual activations—consistently high precision, particularly for rare tactical patterns, demonstrates the validity of our interpretation.
  }
  \label{tab:feature_validation}

  \resizebox{\textwidth}{!}{
    \scriptsize
    \begin{tabular}{l l l c c}
      \toprule
      \textbf{Feature ID} &
      \textbf{Interpretation} &
      \textbf{Verification Rule} &
      \textbf{Precision} &
      \textbf{Recall} \\
      \midrule

      \rowcolor{DetectionBG}
      Tc.0.5593 & [Det]Opp.Knight & Act. @ Opp.Knight & 100\% & 100\% \\
      \rowcolor{DetectionBG}
      \textbf{Tc.0.12663} & [Det]OwnPawn & Act. @ OwnPawn & 100\% & 100\% \\
      \rowcolor{DetectionBG}
      Lorsa.0.13740 & [Det]Opp.Queen $\leftarrow$ OwnQueen
      & Act. @ Opp.Queen $\leftarrow$ OwnQueen & 98.5\% & 100\% \\
      \midrule

      \rowcolor{SourceTargetSquareBG}
      Tc.14.120 & [Src]Source square& Act. @ Top-1 predicted source & 88.0\% & 1.7\% \\
      \rowcolor{SourceTargetSquareBG}
      \textbf{Tc.14.1214} & [Tgt]Target square & Act. @ Top-1 predicted target & 99.2\% & 78.7\% \\
      \midrule

      \rowcolor{ValueBG}
      Tc.14.8947 & [Val]High Value
      & Act. @ Every squares on chessboard positions predicted value $> 0.8$ & 89.9\% & 41.6\% \\
      \rowcolor{ValueBG}
      Tc.14.4046 & [Val]Low Value
      & Act. @ Every squares on chessboard positions predicted value $<- 0.8$ & 83.1\% & 77.7\% \\
      \midrule
      \rowcolor{CaptureBG}
      \textbf{Tc.5.11039} & [Cap]Queen-exchange
      & Act. @ Queen xchg with Opp.Queen & 74.6\% & 49.5\% \\
      \rowcolor{CaptureBG}
      \textbf{Tc.6.4984} & [Cap]Opp.Piece Captured by OwnPawn
      & Act. @ OwnPawn x Opp.Piece & 97.0\% & 86.6\% \\
      \midrule

      \rowcolor{ComplexTacticBG}
      Tc.3.13116 & [Tac]Check-forced rook/queen loss
      & Act. @ Skewer on Own Rook/Queen & 71.5\% & 11.6\% \\
      \rowcolor{ComplexTacticBG}
      \textbf{Tc.1.6695} & [Tac]Opp.Queen checkmate threats
      & Act. @ Opp.Queen Mate-in-1 Threat target & 67.0\% & 10.0\% \\
      \rowcolor{ComplexTacticBG}
      \textbf{Tc.13.7425} & [Tac]OwnQueen/Rook checkmate target
      & Act. @ Check target by OwnQueen/Rook & 64.5\% & 75.2\% \\
      \rowcolor{ComplexTacticBG}
      \textbf{Tc.13.10331} & [Tac]Eat Opp.Pawn and check target/capture queen
      & Act. @ Opp.Pawn, Check Target/Capture Queen & 74.0\% & 7.7\% \\
      \rowcolor{ComplexTacticBG}
      Lorsa.1.15014 & [Tac]Queen check target surronding Opp.King
      & Act. @ Check target, Adj. at Opp.King & 83.5\% & 33.1\% \\
      \rowcolor{ComplexTacticBG}
      \textbf{Lorsa.8.16275} & [Tac]Opp.Rook/Queen semi-open file check threat
      & Act. @ Check target of Opp.Rook/Queen, 2-move reachability, $\le $2 blockers & 98.3\% & 39.5\% \\
      \midrule

      \rowcolor{SpaceBG}
      Lorsa.7.14439 & [Spa]Squares with the same color as OwnKing
      & Act. @ Squares in with the same color as OwnKing & 100\%  & 8.6\% \\
      \rowcolor{SpaceBG}
      \textbf{Lorsa.0.7083} & [Spa]Opp.Rook rank-wise defensive coverage
      & Act. @ Squares in rank-wise defensive coverage Opp.Knight & 99.3\% & 23.8\% \\
      \midrule
      \rowcolor{PieceMovingBG}
      Lorsa.1.3938 & [Mov]OwnKing movement
      & Act. @ Adj.of OwnKing $\leftarrow$ OwnKing & 100\% & 54.9\% \\
      \rowcolor{PieceMovingBG}
      Lorsa.5.5460 & [Mov]OwnKnight movement
      & Act. @ Knight-reach squares $\leftarrow$ OwnKnight & 100.0\% & 61.3\% \\
      \rowcolor{PieceMovingBG}
      \textbf{Lorsa.7.14502} & [Mov]OwnBishop movement
      & Act. @ Bishop-reach squares $\leftarrow$ OwnBishop & 86.3\% & 65.8\% \\
      \rowcolor{PieceMovingBG}
      \textbf{Lorsa.8.9586} & [Mov]OwnBishop movement
      & Act. @ Bishop-reach squares $\leftarrow$ OwnBishop & 98.7\% & 85.8\% \\
      \bottomrule
    \end{tabular}
  }
\end{table*}

\paragraph{Human Assessment Protocols.}
To evaluate feature interpretability, three human chess players independently scored features for \emph{activation consistency} and \emph{complexity} on a 1-5 scale \cite{circuits_updates_june_2024} across Transcoder features, Lorsa features, and MLP neurons, as shown in Table~\ref{tab:feature_human_eval}.  
For a blind comparison with Transcoder features and MLP neurons, Lorsa features were presented without the $z$-pattern, though observing the full pattern would in practice provide stronger interpretability.

We do not directly compare Lorsa features with MHSA \emph{query-key} patterns~\cite{lczero2024transformer, jenner2024evidence, monroe2024mastering}, because Lorsa activations reflect integrated \emph{query-key-value interactions} rather than \emph{query-key} patterns alone.

\section{Constructing Reasoning Pathways}

\label{sec:discovering_reasoning_pathways}

\begin{algorithm}[t]
\caption{Constructing Reasoning Pathways}
\label{alg:reasoning_path_generation}
\begin{algorithmic}[1]

\REQUIRE Move $m$ of a position; activated feature set $F$; 
feature-to-output steering factor $\alpha$; 
feature-to-feature steering factor $\beta$

\ENSURE Significant feature set $\mathcal{V}_m$, reasoning pathway $\mathcal{G}_m$

\STATE \textbf{Feature selection and pruning}
\STATE Compute $\{\mathrm{Infl}(f,m;\alpha) \mid f \in F\}$
\STATE $\mathcal{V}_m \gets \operatorname{Prune}_{\text{feature}}
       \bigl(\{ f \in F \mid \mathrm{Infl}(f,m;\alpha) \}\bigr)$

\STATE Initialize directed graph $\mathcal{G}_m = (\mathcal{V}_m, \emptyset)$

\STATE \textbf{Edge construction}
\FOR{each ordered pair $(f_i,f_j) \in \mathcal{V}_m \times \mathcal{V}_m$ with $\ell(f_i) < \ell(f_j)$}
    \STATE $w_{ij} \gets \mathrm{Infl}(f_i,f_j;\beta)$
    \IF{$w_{ij} \neq 0$}
        \STATE Add directed edge $f_i \xrightarrow{w_{ij}} f_j$ to $\mathcal{G}_m$
    \ENDIF
\ENDFOR

\STATE \textbf{Edge pruning}
\STATE $\mathcal{G}_m \gets \operatorname{Prune}_{\text{edge}}(\mathcal{G}_m)$

\STATE \textbf{return }
$\{ (\mathcal{V}_m, \mathcal{G}_m) \}_{m=1}^K$

\end{algorithmic}
\end{algorithm}

While SAEs and Transcoders have advanced \emph{circuit analysis} in transformers~\cite{marks2024sparse, olah2020zoom}, omitting attention hinders a complete understanding of cross-token interactions. We therefore apply Lorsas to sparsify attention modules and Transcoders for MLPs, yielding unified sparse replacement layers of the entire transformer. This enables the construction of reasoning pathways, a full-model circuit representation in which nodes are sparse features and edges capture their interactions in producing the targeted model's output(a move). Formally, a reasoning pathway for move $m$ is denoted by $\mathcal{G}_m$ in Algorithm~\ref{alg:reasoning_path_generation}.

Building on feature steering~\cite{bricken2023monosemanticity, ge2024hierattr, chalnev2024improving, heimersheim2024use}, we apply a unified framework to steer both Transcoder and Lorsa features. By manipulating features to construct patched activations~\cite{nanda22attribution, wang23ioi, zhang24patching} and observing the resulting changes, we measure their causal effect on predicted move probabilities and downstream feature activations.

\paragraph{Feature Steering.}
A \emph{steered forward pass} intervenes by adding a feature’s decoder direction to the residual stream at layer $l$, scaled by a steering factor $\alpha$:
\begin{equation}
\tilde{h}^l = h^l + \alpha \,(a^l \cdot W_i),
\end{equation}
after which, the forward pass proceeds normally. Here, $l$ indexes computation stages, defined as $l = 2k + \mathbb{I}_{\text{MLP}}$, where $k$ is the transformer block index and $\mathbb{I}_{\text{MLP}} \in \{0,1\}$ indicates whether the intervention is applied at the attention or MLP sublayer.
We characterize downstream effects of feature steering using two notions of indirect effect~\cite{pearl2022direct}: \textbf{feature-to-output effect} and \textbf{feature-to-feature effect}.

\paragraph{Feature-to-output effect.}
For a target output move $m$, we define the \emph{feature-to–output effect} of a feature $f$ as the change in the model-normalized probability $P(m)$ induced by steering $f$ at layer $l$ with the steering factor $\alpha$.
\begin{equation}
\mathrm{Eff}(f, m; \alpha)
= P(m)\big|_{\tilde{h}^l}
- P(m)\big|_{h^l}.
\label{eq:prob_effect}
\end{equation}
This metric directly measures how a feature-level intervention influences the model’s output.

\paragraph{Feature-to-feature effect.}
For a downstream feature $f_j$ at layer $l' > l$, we define the \emph{feature-to-feature effect} as the relative change in its activation $a_j$ under steering:
\begin{equation}
\mathrm{Eff}(f_i, f_j; \alpha)
= \frac{a_j \big|_{h^l \leftarrow \tilde{h}^l} - a_j \big|_{h^l}}{a_j \big|_{h^l}},
\label{eq:feature_effect}
\end{equation}
where $a_j$ denotes the activation of $f_j$. This quantifies how steering an upstream feature propagates causally to downstream features as a proportion of the original activation.

Algorithm~\ref{alg:reasoning_path_generation} describes the construction of reasoning pathways. Specifically, we first select significant features as nodes based on their influence on the output probability, then connect them using feature interactions and prune the resulting graph to a manageable size for analysis. In our paper, we group semantically related features into supernodes~\cite{ameisen2025circuit} for visualization.

In practice, we use the steering factor $\alpha=-1, \beta=-1$, which is consistent with the intuition of zero ablation~\cite{farrell2024applying} of the feature. We discuss in Appendix~\ref{appendix:steering_factor} that this choice is not only intuitively but also empirically reasonable. In Algorithm~\ref{alg:reasoning_path_generation}, node- and edge-level pruning shown in Appendix~\ref{appendix:reasoning_path_generation}~\cite{ameisen2025circuit} is applied to enhance visual clarity.

\section{Learning from Reasoning Pathways}
\label{sec:case_study}

By analyzing \textbf{reasoning pathways}, we derive \textbf{mechanistic findings} about LC0's \textbf{internal computation} from inputs to policy outputs. In this section, we focus on one representative case shown in Figure~\ref{fig:case_compilation}, while providing additional cases in Appendix~\ref{appendix:case_supernode}. This case study is intended to make LC0's internal computation concrete and visually interpretable, and to show how reasoning pathways can reveal both verifiable tactical computations and a possible computational source of model blindspots.

In the position shown in Figure~\ref{fig:case_compilation}(\textbf{A}, left), the optimal move for White is \textcolor{BestMove}{\textbf{\texttt{Qxh7+}}}, which delivers checkmate in one move. However, \textbf{LC0} assigns comparable probabilities to three candidate moves {\textbf{\texttt{fxg7+}}} (35.8\%), \textcolor{BestMove}{\textbf{\texttt{Qxh7+}}} (28.8\%), and \textcolor{Alternative}{\textbf{\texttt{Qxd7}}} (26.1\%). Figure~\ref{fig:case_compilation}(\textbf{A}, right) visualizes the reasoning pathways of these candidate moves.

The resulting reasoning pathways reveal three mechanistic findings about the model's internal computation. First, LC0 transfers bishop-control information from \textbf{\texttt{d3}}, exhibiting cross-layer superposition (\textbf{\emph{Finding 1}}). Second, the model incorporates the defensive coverage of the opponent's rook on \textbf{\texttt{f7}} when assessing the same move (\textbf{\emph{Finding 2}}). Third, the model's ambiguity in this position can be traced to an over-evaluation of defensive dependencies associated with the \textbf{\texttt{g2}} pawn, which biases LC0 toward a more conservative continuation (\textbf{\emph{Finding 3}}). Findings~1 and~2 illustrate tactically meaningful and causally verifiable internal computation, while Finding~3 suggests a possible computational mechanism underlying LC0's blindspots. We detail these findings in the following.

\begin{figure*}[t]
    \centering
    \includegraphics[width=0.99\linewidth]{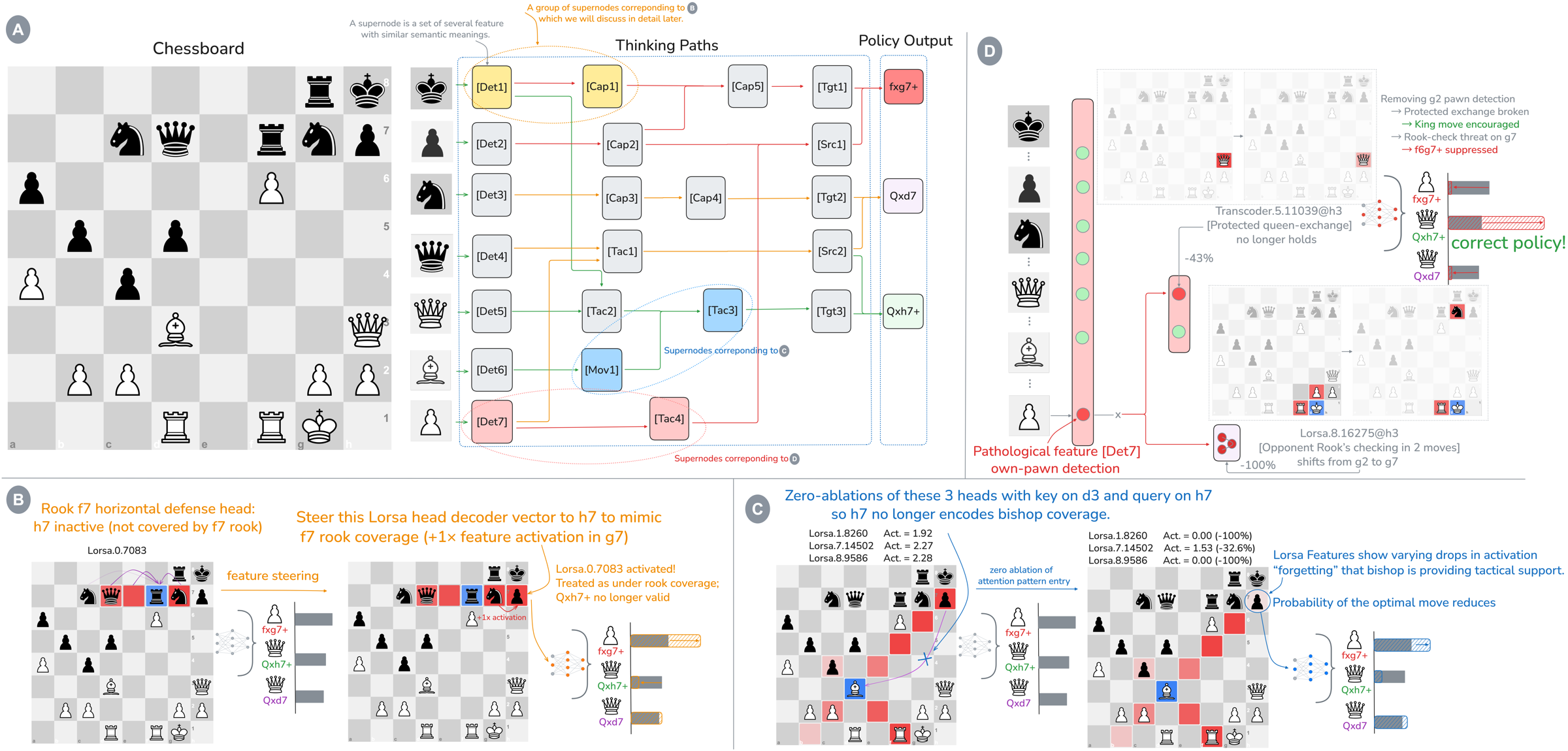}
    \caption{A complete case study with illustration highlighting the key tactical structure discussed in this section.
    (A) Input position and extracted reasoning pathways.
    (B) Finding 1: Mechanistic validation of bishop-movement features that transfer bishop coverage information via targeted attention pattern masking.
    (C) Finding 2: Copying activation of \textbf{\texttt{Lorsa.0.7083}} opponent’s rook coverage feature from \textbf{\texttt{g7}} to \textbf{\texttt{h7}} alters perceived rook defense and thus suppresses \textbf{\texttt{Qxh7+}}.
    (D) Finding 3: The model’s detection of the \textbf{\texttt{g2}} pawn triggers perceived defensive stability, reducing sensitivity to forced-win opportunities.
    }
    \label{fig:case_compilation}
\end{figure*}

\par
\smallskip
\noindent
\textbf{\emph{Finding 1: Bishop-control information is transferred via attention modules, exhibiting cross-layer superposition.}}
\smallskip
\par

The queen-checkmate relies on \textbf{\texttt{d3}} bishop's control over \textbf{\texttt{h7}}. We identify three Lorsa features activated on \textbf{\texttt{h7}} that receive contributions from \textbf{\texttt{d3}}, as shown in Figure~\ref{fig:case_compilation}\textbf{B} (visualized in Figure~\ref{fig:finding1_features}).
These features aggregate signals from \textbf{\texttt{d3}} bishop square and transfer the bishop’s control over \textbf{\texttt{h7}} in layer 1, 7, and 8. 

To verify the functions of these three Lorsa features, we zero-ablated the attention entries from key position \textbf{\texttt{d3}} to query position \textbf{\texttt{h7}} for all three Lorsa features. This sharply reduced their activation at \textbf{\texttt{h7}} (Figure~\ref{fig:case_compilation}\textbf{B}, right) and lowered the probability of \textcolor{BestMove}{\textbf{\texttt{Qxh7+}}} from 28.75\% to 23.35\%.

Features with similar functionality may be distributed across layers, and ablating any single one only weakly affects the output, 
reflecting a form of cross-layer superposition~\cite{mcgrath2023hydra, lindsey2024crosscoder}. Applying a stronger steering factor ($\alpha = -2$) to each decoder vector $W_{D,i}$ in the residual stream results in a near-total collapse, with the probability of \textcolor{BestMove}{\textbf{\texttt{Qxh7+}}} dropping from 28.75\% to 0.7\%.

This indicates that reasoning pathways reveal how the model transfers piece control information over the board with attention modules, which is leveraged for coordinated attacks, and that such features are organized via cross-layer superposition
(validated in Appendix~\ref{appendix:cl_superposition}).

\par
\smallskip
\noindent
\textbf{\textit{Finding 2: LC0 incorporates the defensive coverage of the opponent's rook.}}
\smallskip
\par

Since the black \textbf{\texttt{f7}} rook does not defend \textbf{\texttt{h7}}, \textcolor{BestMove}{\textbf{\texttt{Qxh7+}}} delivers checkmate. We identify \textbf{\texttt{Lorsa.0.7083@}} (\emph{opponent's rook coverage}) as encoding information about the opponent’s rook defensive coverage. As illustrated in Figure~\ref{fig:case_compilation}(\textbf{C}) and Figure~\ref{fig:finding2_features}(1), this feature is activated on squares in rank-wise defensive coverage by an opponent's rook, which has been verified in Table~\ref{tab:feature_validation}.

The \emph{opponent's rook coverage} feature is activated on the \textbf{\texttt{g7}} knight rather than the \textbf{\texttt{h7}} pawn, signaling that the pawn is outside the opponent's rook’s defensive coverage. 

By copying the activation of the \emph{opponent's rook coverage} feature from \textbf{\texttt{g7}} to \textbf{\texttt{h7}}, we simulate the model perceiving the \textbf{\texttt{h7}} pawn as defended. This suppresses downstream \emph{attack with queen} feature activations (Table~\ref{tab:rook_defense_feature_causal_chain}), neutralizing the diagonal-assisted checkmate (Appendix~\ref{appendix:case_features}) and finally reducing the probability of \textcolor{BestMove}{\textbf{\texttt{Qxh7+}}} from $28.75\%$ to $9.32\%$. This demonstrates that if the \textbf{\texttt{h7}} pawn were protected by the opponent's \textbf{\texttt{f7}} rook, \textcolor{BestMove}{\textbf{\texttt{Qxh7+}}} would no longer be a viable move. Overall, this study suggests that reasoning pathways can reveal how the model incorporates information about the opponent's defensive structure to guide its final decision.

\par
\smallskip
\noindent
\textbf{\emph{Finding 3: The ambiguity in this position can be traced to an over-evaluation of defensive dependencies.}}
\smallskip
\par

\begin{table}[t]
\centering
\renewcommand{\arraystretch}{1.5}
\caption{Causal effects of steering by injecting the opponent’s rook coverage feature \textbf{\texttt{Lorsa.0.7083}} into \textbf{\texttt{h7}}. After this intervention, the model perceives \textbf{\texttt{h7}} as being defended by the black rook on \textbf{\texttt{f7}}, which suppresses downstream features corresponding to related tactics.}
\label{tab:rook_defense_feature_causal_chain}
\scriptsize
\setlength{\tabcolsep}{4pt} 
\begin{tabular}{
p{0.30\linewidth}
p{0.15\linewidth}
p{0.43\linewidth}
}
\toprule
\textbf{Feature} & \textbf{$\Delta$ Act.} & \textbf{Interpretation} \\
\midrule
\multicolumn{3}{l}{\textbf{Steered Feature}} \\
\textbf{\texttt{Lorsa.0.7083@h7}} & +1x (copy from \textbf{\texttt{g7}}) & [Spa] Opponent Rook rank-wise defensive coverage \\
\midrule
\multicolumn{3}{l}{\textbf{Downstream Features (affected by steering)}} \\
\textbf{\texttt{Lorsa.1.9198@h7}} & $\downarrow 100\%$ & Opponent knight blocks rook protection \\
\textbf{\texttt{Lorsa.1.15014@h7}} & $\downarrow 100\%$ & Diagonal-assisted queen checking threats near opponent king \\
\textbf{\texttt{Lorsa.1.16036@h7}} & $\downarrow 100\%$ & Multi-piece attack--defense around king-front pawns \\
\textbf{\texttt{Tc.8.9724@h7}} & $\downarrow 100\%$ & Queen target \\
\textbf{\texttt{Tc.13.10331@h7}} & $\downarrow 14.8\%$ & Eat Opp.Pawn and check target \\
\textbf{\texttt{Tc.13.7425@h7}} & $\downarrow 31.5\%$ & OwnQueen checkmate target \\
\bottomrule
\end{tabular}
\end{table}

Revisiting the position, the model's decision is initially ambiguous. Removing the player's own \textbf{\texttt{g2}} pawn, which is seemingly unrelated to the one-move checkmate, causes the model to commit to \textcolor{BestMove}{\textbf{\texttt{Qxh7+}}}. Similarly, zero ablation ($\alpha=-1$) of feature \textbf{\texttt{Tc.0.12663@g2}}—an \emph{own-pawn-detection} feature with 100\% precision and accuracy (Figure~\ref{fig:finding3_features})—produces a similar effect, as shown in Table~\ref{tab:g2_vs_feature_removal}.

Specifically, ablating the \emph{own-pawn-detection} feature reduces the activation of \textbf{\texttt{Tc.10.1958@h3}} (\emph{queen-exchange}) by 37\%. Since this feature strongly correlates with whether the queen is protected (Figure~\ref{fig:tc_10_1958_protected}), the suppression suggests that the model perceives the queen on \textbf{\texttt{h3}} as unprotected and thus favors moving the queen. To confirm this causal link, we manually steered the \emph{queen-exchange} feature ($\alpha = -2$), which increased the probabilities of the attacking continuations \textcolor{BestMove}{\textbf{\texttt{Qxh7+}}} and \textcolor{Alternative}{\textbf{\texttt{Qxd7}}} by 3.62\% and 2.03\%, respectively.

The model initially perceives a check threat from the opponent's rook at \textbf{\texttt{g2}}. After ablating the \emph{own-pawn-detection} feature on \textbf{\texttt{g2}}, the square in front of the king becomes empty, creating a potential threat at \textbf{\texttt{g7}}. Consequently, the activation of \textbf{\texttt{Lorsa.8.16275@g2}}—encoding opponent rook or queen check threats within two moves and $\le $2 blockers—drops to zero and shifts to the \textbf{\texttt{g7}} knight square, indicating that the model re-evaluates risks and treats \textbf{\texttt{g7}} as newly vulnerable.

To further validate the feature's causal role, we copy the original \textbf{\texttt{g2}} activation to \textbf{\texttt{g7}} and run a forward pass on the original board. This intervention reduces the probability of \textcolor{Suboptimal}{\textbf{\texttt{fxg7+}}} from $35.78\%$ to $32.58\%$, showing that when \textbf{\texttt{g7}} is perceived as vulnerable to a rook attack, the model becomes less inclined to move the defending \textbf{\texttt{f6}} pawn. Together, these results establish a causal chain in which the over-valuation of defensive constraints can place the model in a state of strategic complacency, reducing its sensitivity to immediate tactical wins.

Our findings show that features and reasoning pathways reveal LC0’s internal decision-making; all identified features are validated in Table~\ref{tab:feature_validation} and visualized in Appendix~\ref{appendix:case_features}.

{\setlength{\textfloatsep}{6pt}
\begin{table}
\centering
\caption{Top-3 move logit/probability distributions for the original position, the position without the g2 pawn, and the intervention that ablates \textbf{Tc.0.12663@14}.}
\label{tab:g2_vs_feature_removal}
\small
\begin{tabular*}{\linewidth}{@{\extracolsep{\fill}} lccc}
\toprule
\textbf{Move} 
& \textbf{Orig.} 
& \textbf{w/o g2} 
& \textbf{w/o Tc.0.12663@14} \\
\midrule
\textcolor{BestMove}{\textbf{\texttt{Qxh7+}}} 
& 2.04 / 28.8\% 
& \textbf{3.59 / 91.7\%} 
& \textbf{3.53 / 90.6\%} \\

\textcolor{Suboptimal}{\textbf{\texttt{fxg7+}}}
& \textbf{2.26 / 35.8\%} 
& -2.29 / 0.26\% 
& -0.79 / 1.20\% \\

\textcolor{Alternative}{\textbf{\texttt{Qxd7}}}  
& 1.95 / 26.1\% 
& -1.42 / 0.61\%
& -2.24 / 0.28\% \\
\bottomrule
\end{tabular*}
\end{table}
\section{Quantitative Results}
\label{sec:dive_into_thinking_paths}
From the case study in Sec.~\ref{sec:case_study}, we observe that different candidate moves are primarily driven by disjoint features, and that in deeper layers, these features progressively converge on the corresponding source and target squares. To validate these observations, we perform the following experiments.

\subsection{Distribution and Parallelism Evaluation.}
\label{sec:parallelism}
We first construct a \emph{universal evaluation dataset} stratified by model confidence, defined as the \textbf{probability margin} between the top-two moves. The dataset contains four subsets: \emph{All}, \emph{Confident}, \emph{Confused}, and \emph{Same-source} (Appendix~\ref{appendix:universal_datasets} for the detailed protocol). We compare path-level metrics for pathways corresponding to top-2 moves of each position input, while \emph{Same-source} means the top-2 policy moves have the same source square (start point of a move).

\paragraph{Path Overlap.} We compute the mean Jaccard similarity~\cite{jaccard1912distribution} between the significant feature sets $\mathcal{V}_i$ and $\mathcal{V}_j$ of two policy outputs, pruning each set to the top 100 most influential features for standardized comparison.

The \emph{Path Overlap} $O$ is defined as:
\begin{equation}
O(\mathcal{V}_i, \mathcal{V}_j)
=
\frac{|\mathcal{V}_i \cap \mathcal{V}_j|}{|\mathcal{V}_i \cup \mathcal{V}_j|}.
\end{equation}
Table~\ref{tab:path_overlap} shows that top-2 path overlaps remain below 15\%, indicating a highly distributed reasoning regime in which different moves utilize largely disjoint feature sets. 
The \textbf{All} subset measures overlaps between top-2 moves from random real positions (7.05\%), while the \textbf{Baseline} reports the expected overlap between randomly sampled features (0.17\%), suggesting that LC0 retains limited but non-trivial global shared representations.

The contrast between \emph{Confident} (1.63\%) and \emph{Confused} (8.39\%) subsets reveals that higher decision certainty aligns with greater feature exclusivity. Maximal overlap in the \emph{Same-source} indicates top-2 moves will reuse the representation of the shared source square.

\begin{table}[t]
\centering
\caption{Path overlaps under different evaluation settings, with values close to zero, suggesting that different moves arise from largely disjoint significant features.}
\label{tab:path_overlap}
\scriptsize
\resizebox{0.99\linewidth}{!}{%
\begin{tabular}{lccccc}
\toprule
\textbf{Subset} & All & Confident & Confused & \makecell{Same-\\source} & baseline\\
\midrule
\textbf{Mean path overlap} & 7.05\% & 1.63\% & 8.39\% & \textbf{13.7\%} & 0.17\%\\
\bottomrule
\end{tabular}%
}
\end{table}

\paragraph{Path Cohesion and Path Coupling.} 
From the spatial and causal relations between two reasoning pathways, we propose \emph{path cohesion} and \emph{path coupling} to quantify intra-pathway causal effects and inter-pathway causal effects. For these metrics, we set the steering factor $\alpha = -1$, representing a complete feature ablation of upstream feature's contribution to downstream activations.

\textbf{Path Cohesion} ($\mathcal{C}_{oh}$) of a reasoning pathway $\mathcal{V}_m$ is defined as the average \emph{feature-to-feature effect} by ordered feature pairs along the pathway:
\begin{equation}
\mathcal{C}_{oh}(\mathcal{V}_m)
\;=\;
\frac{1}{|\mathcal{E}_m|}
\sum_{\substack{
f_i, f_j \in \mathcal{V}_m \\
\ell(f_i) < \ell(f_j)
}}
\mathrm{Eff}(f_i, f_j; -1),
\end{equation}
where $|\mathcal{E}_m|$ denotes the number of ordered feature pairs in the set. 

\textbf{Path Coupling} ($\mathcal{C}_{up}$) measures the average causal influence by feature pairs from two reasoning pathways, where each pathway is represented by its significant feature set $\mathcal{V}$:
\begin{equation}
\mathcal{C}_{up}(\mathcal{V}_1, \mathcal{V}_2)
=
\mathbb{E}_{(f_i, f_j) \sim \mathcal{V}_1 \times \mathcal{V}_2}
\Bigl[
\mathrm{Eff}(f_{\mathrm{up}}, f_{\mathrm{down}}; -1\bigr)
\Bigr],
\end{equation}
where $(f_{\mathrm{up}}, f_{\mathrm{down}})$ denotes the ordered pair determined by the model’s causal ordering between $f_i$ and $f_j$. 
As shown in Figure~\ref{fig:coh_cup}, Path Cohesion consistently exceeds Path Coupling, indicating that causal effects within a pathway are stronger than cross-path interventions.

In \emph{Confident} positions, Path Cohesion is high ($67.1\%$) while Path Coupling remains low ($1.8\%$), reflecting strong within-path causal coherence. 
The \emph{Same-source} subset exhibits slightly higher values than \emph{All}, indicating shared origins without loss of move-specific reasoning structure.

\begin{figure}
    \centering
    \includegraphics[width=0.99\linewidth]{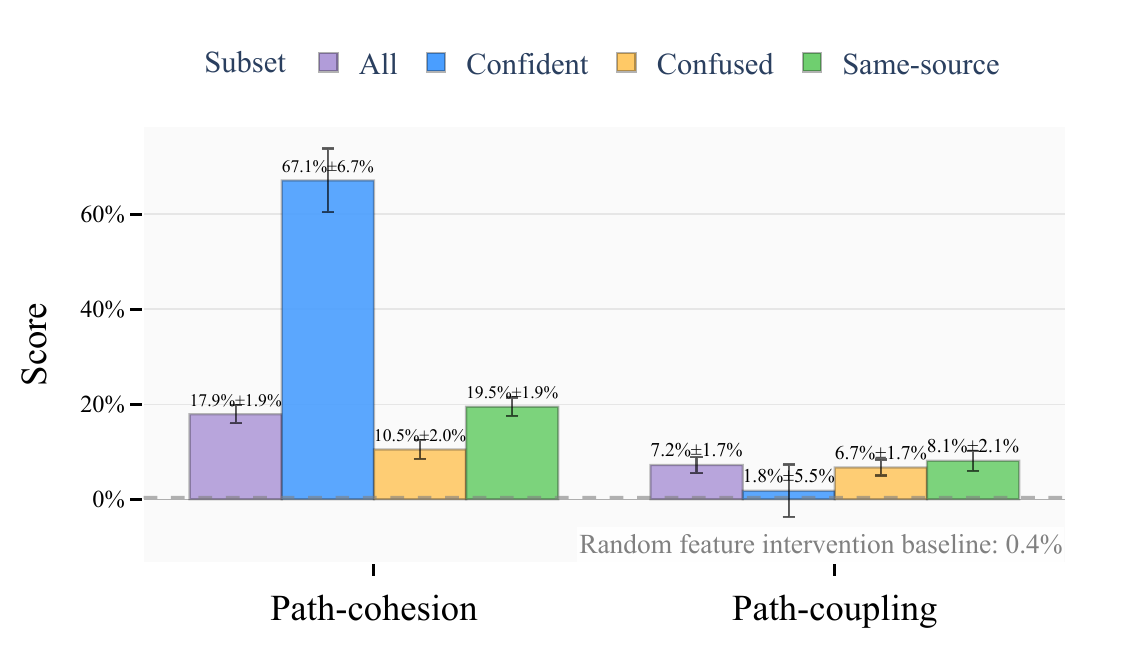}
    \caption{Mean Path Cohesion and Path Coupling between feature sets of top-2 moves across different chess positions in the dataset.}
    \label{fig:coh_cup}
\end{figure}

\begin{figure}[t]
    \centering
    \begin{subfigure}{0.49\linewidth}
        \centering
        \includegraphics[width=\linewidth]{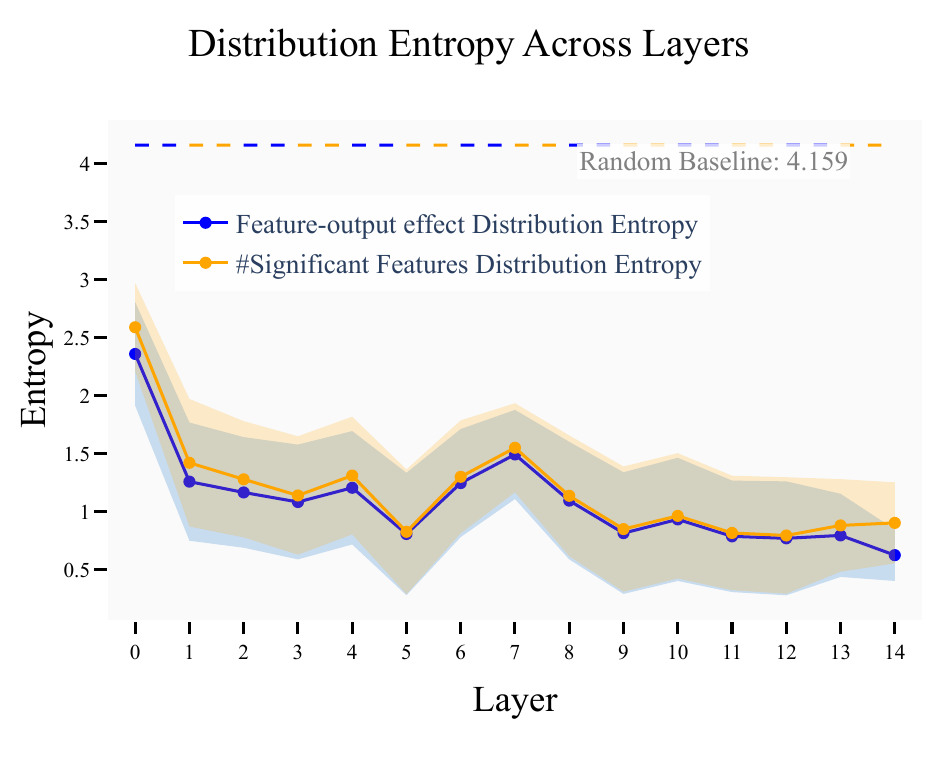}
        \caption{Entropy of significant feature counts and feature-to-output effect distributions across layers. Both exhibit a downward trend, indicating that the quantity and influence of critical features converge in deeper layers.}
        \label{fig:entropy_distribution}
    \end{subfigure}
    \hfill
    \begin{subfigure}{0.49\linewidth}
        \centering
        \includegraphics[width=\linewidth]{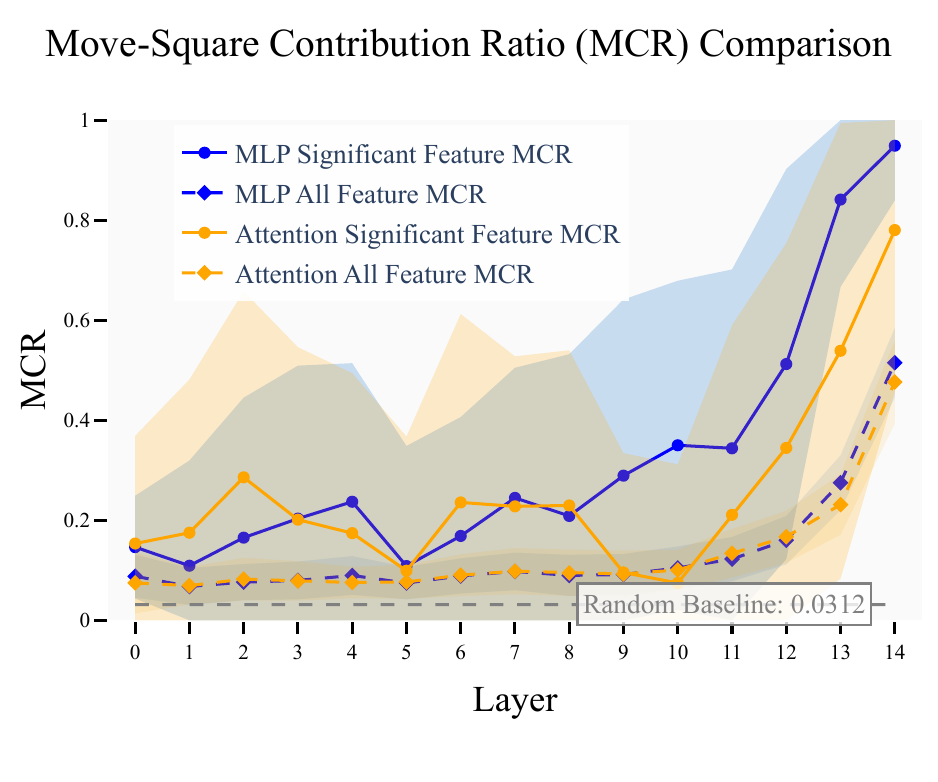}
        \caption{MCR for significant rise layer-wise, nearing 1 by layer 14, with MCR of all features also rising. This suggests a progressive concentration of decision-relevant features on the move-square.}
        \label{fig:spatial_distribution}
    \end{subfigure}

    \caption{Detailed analysis of feature distributions. (a) Entropy distribution across layers, indicating concentration at specific squares. (b) Spatial layout of features on the board.}
    \label{fig:feature_distributions}
\end{figure}

\subsection{Layer-wise Convergence of Feature Contribution}
\label{sec:position_convergence}
We further analyze how reasoning-pathway features become progressively concentrated on the move-relevant squares in deeper layers. To quantify this trend, we first define the \emph{Feature-to-output effect entropy}
\begin{equation}
H_{\mathrm{effect}}
=
-\sum_{i=1}^{n}
\frac{w_i}{\sum_{j=1}^{n} w_j}
\log\!\left(
\frac{w_i}{\sum_{j=1}^{n} w_j}
\right),
\end{equation}
where $w_i$ is the total feature-to-output effect aggregated over significant features at board square $i$. Lower entropy indicates that decision-relevant information is concentrated on fewer squares. We also compute the entropy of the number of significant features across board positions, which measures whether significant features are spatially concentrated or broadly distributed. These results are shown in Figure~\ref{fig:entropy_distribution}.

To directly measure concentration on the move squares, we define the \emph{Move-Square Contribution Ratio} (MCR):
\begin{equation}
\mathrm{MCR}
=
\frac{W_{\mathrm{start}} + E_{\mathrm{end}}}{\sum_{p=1}^{n} W_p},
\qquad
W_p=\sum_f |\mathrm{Eff}(f,p)|,
\end{equation}
where $E_p$ is the total feature-to-output effect aggregated over features at board position $p$. Higher MCR indicates that more move-relevant information is concentrated on the source and target squares. As shown in Figure~\ref{fig:spatial_distribution}, computed over 1{,}000 random positions and their top policy moves, MCR increases with layer depth for both significant and activated features. This is consistent with LC0's attention policy head, where each move logit is computed from the representations of its source and target squares.

Together, these quantitative results provide evidence that LC0 supports parallel reasoning across candidate moves, while progressively aggregating move-specific information onto source and target squares to form the final policy logits.

\section{Related Work}
\label{sec:related_work}
\textbf{Chess as a Testbed for Reasoning.} Chess is as an ideal domain for studying internal reasoning due to its structured rules and multi-step logic. Neural networks have long succeeded in chess, from Deep Blue~\cite{campbell2002deep} to AlphaZero~\cite{silver2018general}. Building on the MCTS-based self-play framework of AlphaZero, modern Transformer models—most notably the Leela Chess Zero (LC0) series~\cite{lczero2024transformer}—have reached grandmaster-level performance in chess.

\textbf{Sparse Decomposition and Replacement Layers.} Since internal representations in neural networks are highly compressed in superposition~\cite{elhage2022tms, jermyn2024attentionsuperposition}, sparse decomposition methods like Sparse Autoencoders (SAEs) are employed to learn monosemantic features from superposition \cite{bricken2023monosemanticity, karvonen2024measuring}. Transcoders extend this paradigm by decomposing MLP computations and constructing replacement layers for MLP to extract interpretable computational features rather than just static activations~\cite{dunefsky2024transcoder, ge2024hierattr, ameisen2025circuit}. Furthermore, Low-Rank Sparse Attention (Lorsa) has been introduced to resolve the challenge of attention superposition~\cite{he2025towards}. While these methods have primarily been validated in large language models, we apply them here to construct unified replacement layers for both MLP and attention modules.

\textbf{Mechanistic Interpretability in Chess.} Prior interpretability studies have identified linearly separable concepts aligned with human knowledge in AlphaZero~\cite{mcgrath2022acquisition}, as well as machine-specific concepts transferable to humans~\cite{schut2025bridging}. Subsequent work used Sparse Autoencoders~(SAEs) to recover high-fidelity board-state properties such as piece locations~\cite{karvonen2024measuring}. \citet{poupart2024contrastive} further applied contrastive SAEs to LC0, analyzing features associated with optimal and suboptimal trajectories. Cross-layer Transcoders in Chessformer~\cite{ameisen2025circuit, anonymous2025chessformer} revealed tactical concepts in a domain-specific transformer. However, these approaches remain primarily static and local, with only specific attention heads shown to support look-ahead reasoning~\cite{jenner2024evidence}. Similar techniques in Othello-GPT attempted patch-free circuit discovery via SAE~\cite{he2024othello}. Our work extends these efforts by moving from static representation analysis to uncovering dynamic, end-to-end computations.

\section{Discussion and Limitations}
\label{sec:discussion_and_limitations}
We use replacement layers—Transcoders for MLPs and Lorsas for attentions—for unsupervised feature discovery. Unlike probing~\cite{mcgrath2022acquisition, jenner2024evidence}, unsupervised dictionary learning more directly exposes the model’s emergent internal representations. This enables the identification of features that do not correspond to predefined human concepts, offering a window into reasoning strategies that may differ from human intuition and exhibit more multi-step structure, and thereby providing insight into how advanced capabilities are represented in LC0.

For example, Figure~\ref{fig:win_or_go_home} shows that when selecting a checkmating move, the model also accounts for whether the move prevents the opponent’s mating threat. Figure~\ref{fig:looking_ahead} not only recovers the \emph{looking ahead} phenomenon reported in prior work~\cite{jenner2024evidence}, but also reveals a more complete chain of computation. Figure~\ref{fig:grandmaster-level_sacrificing} presents a harder middlegame position in which BT4 correctly finds a sacrifice, with a reasoning pathway that reflects several grandmaster-like considerations, though without explicitly encoding the coordination of the knights on \textbf{\texttt{c3}}.

Overall, this work takes an exploratory step toward understanding the full computation of a capable model. By studying a grandmaster-level chess transformer, our results highlights the potential for mechanistic interpretability to shed light on the internal reasoning of highly capable models.

\subsection{Limitation}
We discuss the limitations of our approach. These observations also highlight multiple directions in which future work may further improve upon our current framework.

\paragraph{Generality.} Our study is currently confined to a single architecture, LC0 BT4, which restricts the diversity. In fact, our use of replacement layers and the method of constructing reasoning pathways is theoretically applicable in any transformer models. Generalization across architectures remains to be studied.

\paragraph{Auto-interpretation.} Automatic interpretation remains challenging. Even with structured inputs such as the board position, feature activation locations, and high-$z$ positions for Lorsa features, LLMs still struggle to produce reliable interpretations~\cite{paul2025auto}. This may require more agentic methods, and many features are easier to interpret within a reasoning pathway than in isolation.

\paragraph{Feature Validation, Splitting and Absorption.} Our binary and rule-based validation is inherently conservative: it penalizes low activations on irrelevant squares as false positives, thereby magnifying noise. In practice, we only consider a feature as active on a square if its activation exceeds 10\% of the feature’s maximum activation. Despite this threshold, the precision of $\sim70\%$ for some tactical features is reported in Table~\ref{tab:feature_validation}.
Meanwhile, lower recall for some features likely reflects an overly coarse interpretation, suggesting feature splitting~\cite{chanin2025absorption} or absorption, where the model represents the direction as finer-grained distinctions than our explanatory rules capture.

\paragraph{Uninterpretable features and feature taxonomy.} Some features remain difficult to interpret, such as dense features~\cite{circuits_updates_june_2024}. This may stem from intrinsic model opacity, reconstruction error of replacement layers, features with high-dimensionality~\cite{joshua2025linear}, or our limited ability to understand board-level characteristics. A systematic feature taxonomy could help clarify how many types of features exist, whether clear boundaries separate them, and which characteristicas are associated with lower interpretability.

\section{Conclusion}
We extract interpretable computational features from LC0’s MLP and attention modules using sparse replacement layers. And we introduce a method to construct \emph{reasoning pathways} for any input position via feature steering. Through a detailed case study, we demonstrate that reasoning pathways offer strong interpretability, unveil the reasoning logic within the model, and can be validated through targeted intervention. Finally, our quantitative studies reveal that the model operates through a parallel reasoning mechanism and its converging alignment with the model's architectural inductive biases. Our method and findings provide insights into fully understanding a grandmaster-level chess transformer, LC0, and open new avenues for humans to learn knowledge from a model's internal representations.

\bibliography{example_paper}

@misc{nanda22attribution,
  author       = {Neel Nanda},
  title        = {Attribution Patching: Activation Patching at Industrial Scale},
  howpublished = {\url{https://www.neelnanda.io/mechanistic-interpretability/attribution-patching}},
  year         = {2022},
  bibsource    = {dblp computer science bibliography, https://dblp.org}
}

@inproceedings{paul2025auto,
  author       = {Gon{\c{c}}alo Paulo and
                  Alex Mallen and
                  Caden Juang and
                  Nora Belrose},
  editor       = {Aarti Singh and
                  Maryam Fazel and
                  Daniel Hsu and
                  Simon Lacoste{-}Julien and
                  Felix Berkenkamp and
                  Tegan Maharaj and
                  Kiri Wagstaff and
                  Jerry Zhu},
  title        = {Automatically Interpreting Millions of Features in Large Language
                  Models},
  booktitle    = {Forty-second International Conference on Machine Learning, {ICML}
                  2025, Vancouver, BC, Canada, July 13-19, 2025},
  series       = {Proceedings of Machine Learning Research},
  publisher    = {{PMLR} / OpenReview.net},
  year         = {2025},
  url          = {https://proceedings.mlr.press/v267/paulo25a.html},
  bibsource    = {dblp computer science bibliography, https://dblp.org}
}

@inproceedings{ruoss2024amortized,
  author       = {Anian Ruoss and
                  Gr{\'{e}}goire Del{\'{e}}tang and
                  Sourabh Medapati and
                  Jordi Grau{-}Moya and
                  Kevin Li and
                  Elliot Catt and
                  John Reid and
                  Cannada A. Lewis and
                  Joel Veness and
                  Tim Genewein},
  editor       = {Amir Globersons and
                  Lester Mackey and
                  Danielle Belgrave and
                  Angela Fan and
                  Ulrich Paquet and
                  Jakub M. Tomczak and
                  Cheng Zhang},
  title        = {Amortized Planning with Large-Scale Transformers: {A} Case Study on
                  Chess},
  booktitle    = {Advances in Neural Information Processing Systems 38: Annual Conference
                  on Neural Information Processing Systems 2024, NeurIPS 2024, Vancouver,
                  BC, Canada, December 10 - 15, 2024},
  year         = {2024},
  url          = {http://papers.nips.cc/paper\_files/paper/2024/hash/78f0db30c39c850de728c769f42fc903-Abstract-Conference.html},
  bibsource    = {dblp computer science bibliography, https://dblp.org}
}

@article{silver2016mastering,
  author       = {David Silver and
                  Aja Huang and
                  Chris J. Maddison and
                  Arthur Guez and
                  Laurent Sifre and
                  George van den Driessche and
                  Julian Schrittwieser and
                  Ioannis Antonoglou and
                  Vedavyas Panneershelvam and
                  Marc Lanctot and
                  Sander Dieleman and
                  Dominik Grewe and
                  John Nham and
                  Nal Kalchbrenner and
                  Ilya Sutskever and
                  Timothy P. Lillicrap and
                  Madeleine Leach and
                  Koray Kavukcuoglu and
                  Thore Graepel and
                  Demis Hassabis},
  title        = {Mastering the game of Go with deep neural networks and tree search},
  journal      = {Nat.},
  volume       = {529},
  number       = {7587},
  pages        = {484--489},
  year         = {2016},
  url          = {https://doi.org/10.1038/nature16961},
  doi          = {10.1038/NATURE16961},
  bibsource    = {dblp computer science bibliography, https://dblp.org}
}

@inproceedings{seo2015solving,
  author       = {Min Joon Seo and
                  Hannaneh Hajishirzi and
                  Ali Farhadi and
                  Oren Etzioni and
                  Clint Malcolm},
  editor       = {Llu{\'{\i}}s M{\`{a}}rquez and
                  Chris Callison{-}Burch and
                  Jian Su and
                  Daniele Pighin and
                  Yuval Marton},
  title        = {Solving Geometry Problems: Combining Text and Diagram Interpretation},
  booktitle    = {Proceedings of the 2015 Conference on Empirical Methods in Natural
                  Language Processing, {EMNLP} 2015, Lisbon, Portugal, September 17-21,
                  2015},
  pages        = {1466--1476},
  publisher    = {The Association for Computational Linguistics},
  year         = {2015},
  url          = {https://doi.org/10.18653/v1/d15-1171},
  doi          = {10.18653/V1/D15-1171},
  bibsource    = {dblp computer science bibliography, https://dblp.org}
}

@inproceedings{lightman2023let,
  author       = {Hunter Lightman and
                  Vineet Kosaraju and
                  Yuri Burda and
                  Harrison Edwards and
                  Bowen Baker and
                  Teddy Lee and
                  Jan Leike and
                  John Schulman and
                  Ilya Sutskever and
                  Karl Cobbe},
  title        = {Let's Verify Step by Step},
  booktitle    = {The Twelfth International Conference on Learning Representations,
                  {ICLR} 2024, Vienna, Austria, May 7-11, 2024},
  publisher    = {OpenReview.net},
  year         = {2024},
  url          = {https://openreview.net/forum?id=v8L0pN6EOi},
  bibsource    = {dblp computer science bibliography, https://dblp.org}
}

@article{romera2024mathematical,
  author       = {Bernardino Romera{-}Paredes and
                  Mohammadamin Barekatain and
                  Alexander Novikov and
                  Matej Balog and
                  M. Pawan Kumar and
                  Emilien Dupont and
                  Francisco J. R. Ruiz and
                  Jordan S. Ellenberg and
                  Pengming Wang and
                  Omar Fawzi and
                  Pushmeet Kohli and
                  Alhussein Fawzi},
  title        = {Mathematical discoveries from program search with large language models},
  journal      = {Nat.},
  volume       = {625},
  number       = {7995},
  pages        = {468--475},
  year         = {2024},
  url          = {https://doi.org/10.1038/s41586-023-06924-6},
  doi          = {10.1038/S41586-023-06924-6},
  bibsource    = {dblp computer science bibliography, https://dblp.org}
}

@article{li2022competition,
  author       = {Yujia Li and
                  David H. Choi and
                  Junyoung Chung and
                  Nate Kushman and
                  Julian Schrittwieser and
                  R{\'{e}}mi Leblond and
                  Tom Eccles and
                  James Keeling and
                  Felix Gimeno and
                  Agustin Dal Lago and
                  Thomas Hubert and
                  Peter Choy and
                  Cyprien de Masson d'Autume and
                  Igor Babuschkin and
                  Xinyun Chen and
                  Po{-}Sen Huang and
                  Johannes Welbl and
                  Sven Gowal and
                  Alexey Cherepanov and
                  James Molloy and
                  Daniel J. Mankowitz and
                  Esme Sutherland Robson and
                  Pushmeet Kohli and
                  Nando de Freitas and
                  Koray Kavukcuoglu and
                  Oriol Vinyals},
  title        = {Competition-Level Code Generation with AlphaCode},
  journal      = {CoRR},
  volume       = {abs/2203.07814},
  year         = {2022},
  url          = {https://doi.org/10.48550/arXiv.2203.07814},
  doi          = {10.48550/ARXIV.2203.07814},
  eprinttype    = {arXiv},
  eprint       = {2203.07814},
  bibsource    = {dblp computer science bibliography, https://dblp.org}
}

@article{tesauro1994td,
  author       = {Gerald Tesauro},
  title        = {TD-Gammon, a Self-Teaching Backgammon Program, Achieves Master-Level
                  Play},
  journal      = {Neural Comput.},
  volume       = {6},
  number       = {2},
  pages        = {215--219},
  year         = {1994},
  url          = {https://doi.org/10.1162/neco.1994.6.2.215},
  doi          = {10.1162/NECO.1994.6.2.215},
  bibsource    = {dblp computer science bibliography, https://dblp.org}
}

@article{monro2024mastering,
  author       = {Daniel Monroe and
                  The Leela Chess Zero Team},
  title        = {Mastering Chess with a Transformer Model},
  journal      = {CoRR},
  volume       = {abs/2409.12272},
  year         = {2024},
  url          = {https://doi.org/10.48550/arXiv.2409.12272},
  doi          = {10.48550/ARXIV.2409.12272},
  eprinttype    = {arXiv},
  eprint       = {2409.12272},
  bibsource    = {dblp computer science bibliography, https://dblp.org}
}

@article{silver2018general,
  title={A general reinforcement learning algorithm that masters chess, shogi, and Go through self-play},
  author={Silver, David and Hubert, Thomas and Schrittwieser, Julian and Antonoglou, Ioannis and Lai, Matthew and Guez, Arthur and Lanctot, Marc and Sifre, Laurent and Kumaran, Dharshan and Graepel, Thore and others},
  journal={Science},
  volume={362},
  number={6419},
  pages={1140--1144},
  year={2018},
  publisher={American Association for the Advancement of Science}
}

@article{silver2017mastering,
  author       = {David Silver and
                  Julian Schrittwieser and
                  Karen Simonyan and
                  Ioannis Antonoglou and
                  Aja Huang and
                  Arthur Guez and
                  Thomas Hubert and
                  Lucas Baker and
                  Matthew Lai and
                  Adrian Bolton and
                  Yutian Chen and
                  Timothy P. Lillicrap and
                  Fan Hui and
                  Laurent Sifre and
                  George van den Driessche and
                  Thore Graepel and
                  Demis Hassabis},
  title        = {Mastering the game of Go without human knowledge},
  journal      = {Nat.},
  volume       = {550},
  number       = {7676},
  pages        = {354--359},
  year         = {2017},
  url          = {https://doi.org/10.1038/nature24270},
  doi          = {10.1038/NATURE24270},
  bibsource    = {dblp computer science bibliography, https://dblp.org}
}

@article{campbell2002deep,
  author       = {Murray Campbell and
                  A. Joseph Hoane Jr. and
                  Feng{-}Hsiung Hsu},
  title        = {Deep Blue},
  journal      = {Artif. Intell.},
  volume       = {134},
  number       = {1-2},
  pages        = {57--83},
  year         = {2002},
  url          = {https://doi.org/10.1016/S0004-3702(01)00129-1},
  doi          = {10.1016/S0004-3702(01)00129-1},
  bibsource    = {dblp computer science bibliography, https://dblp.org}
}

@article{mcgrath2022acquisition,
  author       = {Thomas McGrath and
                  Andrei Kapishnikov and
                  Nenad Tomasev and
                  Adam Pearce and
                  Demis Hassabis and
                  Been Kim and
                  Ulrich Paquet and
                  Vladimir Kramnik},
  title        = {Acquisition of Chess Knowledge in AlphaZero},
  journal      = {CoRR},
  volume       = {abs/2111.09259},
  year         = {2021},
  url          = {https://arxiv.org/abs/2111.09259},
  eprinttype    = {arXiv},
  eprint       = {2111.09259},
  bibsource    = {dblp computer science bibliography, https://dblp.org}
}

@article{schut2025bridging,
  author       = {Lisa Schut and
                  Nenad Toma{\v{s}}ev and
                  Thomas McGrath and
                  Demis Hassabis and
                  Ulrich Paquet and
                  Been Kim},
  title        = {Bridging the human--AI knowledge gap through concept discovery and transfer in {AlphaZero}},
  journal      = {Proceedings of the National Academy of Sciences},
  volume       = {122},
  number       = {13},
  pages        = {e2406675122},
  year         = {2025},
  url          = {https://www.pnas.org/doi/10.1073/pnas.2406675122},
  doi          = {10.1073/pnas.2406675122},
  bibsource    = {dblp computer science bibliography, https://dblp.org}
}

@inproceedings{karvonen2024measuring,
  author       = {Adam Karvonen and
                  Benjamin Wright and
                  Can Rager and
                  Rico Angell and
                  Jannik Brinkmann and
                  Logan Smith and
                  Claudio Mayrink Verdun and
                  David Bau and
                  Samuel Marks},
  editor       = {Amir Globersons and
                  Lester Mackey and
                  Danielle Belgrave and
                  Angela Fan and
                  Ulrich Paquet and
                  Jakub M. Tomczak and
                  Cheng Zhang},
  title        = {Measuring Progress in Dictionary Learning for Language Model Interpretability
                  with Board Game Models},
  booktitle    = {Advances in Neural Information Processing Systems 38: Annual Conference
                  on Neural Information Processing Systems 2024, NeurIPS 2024, Vancouver,
                  BC, Canada, December 10 - 15, 2024},
  year         = {2024},
  url          = {http://papers.nips.cc/paper\_files/paper/2024/hash/9736acf007760cc2b47948ae3cf06274-Abstract-Conference.html},
  bibsource    = {dblp computer science bibliography, https://dblp.org}
}

@misc{lczero2024transformer,
  author = {{Leela Chess Zero Team}},
  title = {Transformer Progress},
  howpublished = {\url{https://lczero.org/blog/2024/02/transformer-progress/}},
  year = {2024},
  month = {February},
}

@inproceedings{jenner2024evidence,
  author       = {Erik Jenner and
                  Shreyas Kapur and
                  Vasil Georgiev and
                  Cameron Allen and
                  Scott Emmons and
                  Stuart J. Russell},
  editor       = {Amir Globersons and
                  Lester Mackey and
                  Danielle Belgrave and
                  Angela Fan and
                  Ulrich Paquet and
                  Jakub M. Tomczak and
                  Cheng Zhang},
  title        = {Evidence of Learned Look-Ahead in a Chess-Playing Neural Network},
  booktitle    = {Advances in Neural Information Processing Systems 38: Annual Conference
                  on Neural Information Processing Systems 2024, NeurIPS 2024, Vancouver,
                  BC, Canada, December 10 - 15, 2024},
  year         = {2024},
  url          = {http://papers.nips.cc/paper\_files/paper/2024/hash/37d9f19150fce07bced2a81fc87d47a6-Abstract-Conference.html},
  bibsource    = {dblp computer science bibliography, https://dblp.org}
}

@article{heimersheim2024use,
  author       = {Stefan Heimersheim and
                  Neel Nanda},
  title        = {How to use and interpret activation patching},
  journal      = {CoRR},
  volume       = {abs/2404.15255},
  year         = {2024},
  url          = {https://doi.org/10.48550/arXiv.2404.15255},
  doi          = {10.48550/ARXIV.2404.15255},
  eprinttype    = {arXiv},
  eprint       = {2404.15255},
  bibsource    = {dblp computer science bibliography, https://dblp.org}
}

@article{he2025towards,
  author       = {Zhengfu He and
                  Junxuan Wang and
                  Rui Lin and
                  Xuyang Ge and
                  Wentao Shu and
                  Qiong Tang and
                  Junping Zhang and
                  Xipeng Qiu},
  title        = {Towards Understanding the Nature of Attention with Low-Rank Sparse
                  Decomposition},
  journal      = {CoRR},
  volume       = {abs/2504.20938},
  year         = {2025},
  url          = {https://doi.org/10.48550/arXiv.2504.20938},
  doi          = {10.48550/ARXIV.2504.20938},
  eprinttype    = {arXiv},
  eprint       = {2504.20938},
  bibsource    = {dblp computer science bibliography, https://dblp.org}
}

@article{kreutz2003dictionary,
  author       = {Kenneth Kreutz{-}Delgado and
                  Joseph F. Murray and
                  Bhaskar D. Rao and
                  Kjersti Engan and
                  Te{-}Won Lee and
                  Terrence J. Sejnowski},
  title        = {Dictionary Learning Algorithms for Sparse Representation},
  journal      = {Neural Comput.},
  volume       = {15},
  number       = {2},
  pages        = {349--396},
  year         = {2003},
  url          = {https://doi.org/10.1162/089976603762552951},
  doi          = {10.1162/089976603762552951},
  bibsource    = {dblp computer science bibliography, https://dblp.org}
}

@article{bricken2023monosemanticity,
    title={Towards Monosemanticity: Decomposing Language Models With Dictionary Learning},
    author={Bricken, Trenton and Templeton, Adly and Batson, Joshua and Chen, Brian and Jermyn, Adam and Conerly, Tom and Turner, Nick and Anil, Cem and Denison, Carson and Askell, Amanda and Lasenby, Robert and Wu, Yifan and Kravec, Shauna and Schiefer, Nicholas and Maxwell, Tim and Joseph, Nicholas and Hatfield-Dodds, Zac and Tamkin, Alex and Nguyen, Karina and McLean, Brayden and Burke, Josiah E and Hume, Tristan and Carter, Shan and Henighan, Tom and Olah, Christopher},
    year={2023},
    journal={Transformer Circuits Thread},
    url={https://transformer-circuits.pub/2023/monosemantic-features/index.html}
}

@article{jaccard1912distribution,
  title={The distribution of the flora in the alpine zone. 1},
  author={Jaccard, Paul},
  journal={New phytologist},
  volume={11},
  number={2},
  pages={37--50},
  year={1912},
  publisher={Wiley Online Library}
}

@article{wang2025attention,
  author       = {Junxuan Wang and
                  Xuyang Ge and
                  Wentao Shu and
                  Zhengfu He and
                  Xipeng Qiu},
  title        = {Attention Layers Add Into Low-Dimensional Residual Subspaces},
  journal      = {CoRR},
  volume       = {abs/2508.16929},
  year         = {2025},
  url          = {https://doi.org/10.48550/arXiv.2508.16929},
  doi          = {10.48550/ARXIV.2508.16929},
  eprinttype    = {arXiv},
  eprint       = {2508.16929},
  bibsource    = {dblp computer science bibliography, https://dblp.org}
}

@article{elhage2022tms,
   title={Toy Models of Superposition},
   author={Elhage, Nelson and Hume, Tristan and Olsson, Catherine and Schiefer, Nicholas and Henighan, Tom and Kravec, Shauna and Hatfield-Dodds, Zac and Lasenby, Robert and Drain, Dawn and Chen, Carol and Grosse, Roger and McCandlish, Sam and Kaplan, Jared and Amodei, Dario and Wattenberg, Martin and Olah, Christopher},
   year={2022},
   journal={Transformer Circuits Thread},
   url={https://transformer-circuits.pub/2022/toy\_model/index.html}
}

@article{jermyn2024attentionsuperposition,
   title={Circuits Updates - January 2024},
   author={Adam Jermyn and Chris Olah and Tom Conerly},
   year={2024},
   journal={Transformer Circuits Thread},
   url={https://transformer-circuits.pub/2024/jan-update/index.html#attn-superposition}
}

@article{he2024othello,
  author       = {Zhengfu He and
                  Xuyang Ge and
                  Qiong Tang and
                  Tianxiang Sun and
                  Qinyuan Cheng and
                  Xipeng Qiu},
  title        = {Dictionary Learning Improves Patch-Free Circuit Discovery in Mechanistic
                  Interpretability: {A} Case Study on Othello-GPT},
  journal      = {CoRR},
  volume       = {abs/2402.12201},
  year         = {2024},
  url          = {https://doi.org/10.48550/arXiv.2402.12201},
  doi          = {10.48550/ARXIV.2402.12201},
  eprinttype    = {arXiv},
  eprint       = {2402.12201},
  bibsource    = {dblp computer science bibliography, https://dblp.org}
}

@inproceedings{zhang24patching,
  author       = {Fred Zhang and
                  Neel Nanda},
  title        = {Towards Best Practices of Activation Patching in Language Models:
                  Metrics and Methods},
  booktitle    = {The Twelfth International Conference on Learning Representations,
                  {ICLR} 2024, Vienna, Austria, May 7-11, 2024},
  publisher    = {OpenReview.net},
  year         = {2024},
  url          = {https://openreview.net/forum?id=Hf17y6u9BC},
  bibsource    = {dblp computer science bibliography, https://dblp.org}
}

@misc{circuits_updates_june_2024,
  title        = {Circuits Updates -- June 2024},
  author       = {{Anthropic Interpretability Team}},
  howpublished = {\url{https://transformer-circuits.pub/2024/june-update/index.html}},
  year         = {2024}
}

@inproceedings{wang23ioi,
  author       = {Kevin Ro Wang and
                  Alexandre Variengien and
                  Arthur Conmy and
                  Buck Shlegeris and
                  Jacob Steinhardt},
  title        = {Interpretability in the Wild: a Circuit for Indirect Object Identification
                  in {GPT-2} Small},
  booktitle    = {The Eleventh International Conference on Learning Representations,
                  {ICLR} 2023, Kigali, Rwanda, May 1-5, 2023},
  publisher    = {OpenReview.net},
  year         = {2023},
  url          = {https://openreview.net/forum?id=NpsVSN6o4ul},
  bibsource    = {dblp computer science bibliography, https://dblp.org}
}

@inproceedings{voita2019analyzing,
  author       = {Elena Voita and
                  David Talbot and
                  Fedor Moiseev and
                  Rico Sennrich and
                  Ivan Titov},
  editor       = {Anna Korhonen and
                  David R. Traum and
                  Llu{\'{\i}}s M{\`{a}}rquez},
  title        = {Analyzing Multi-Head Self-Attention: Specialized Heads Do the Heavy
                  Lifting, the Rest Can Be Pruned},
  booktitle    = {Proceedings of the 57th Conference of the Association for Computational
                  Linguistics, {ACL} 2019, Florence, Italy, July 28- August 2, 2019,
                  Volume 1: Long Papers},
  pages        = {5797--5808},
  publisher    = {Association for Computational Linguistics},
  year         = {2019},
  url          = {https://doi.org/10.18653/v1/p19-1580},
  doi          = {10.18653/V1/P19-1580},
  bibsource    = {dblp computer science bibliography, https://dblp.org}
}

@article{olah2020zoom,
  author       = {Chris Olah and
                  Nick Cammarata and
                  Ludwig Schubert and
                  Gabriel Goh and
                  Michael Petrov and
                  Shan Carter},
  title        = {Zoom In: An Introduction to Circuits},
  journal      = {Distill},
  volume       = {5},
  number       = {3},
  year         = {2020},
  url          = {https://doi.org/10.23915/distill.00024.001},
  doi          = {10.23915/distill.00024.001},
  bibsource    = {dblp computer science bibliography, https://dblp.org}
}

@inproceedings{gao2024oaisae,
  author       = {Leo Gao and
                  Tom Dupr{\'{e}} la Tour and
                  Henk Tillman and
                  Gabriel Goh and
                  Rajan Troll and
                  Alec Radford and
                  Ilya Sutskever and
                  Jan Leike and
                  Jeffrey Wu},
  title        = {Scaling and evaluating sparse autoencoders},
  booktitle    = {The Thirteenth International Conference on Learning Representations,
                  {ICLR} 2025, Singapore, April 24-28, 2025},
  publisher    = {OpenReview.net},
  year         = {2025},
  url          = {https://openreview.net/forum?id=tcsZt9ZNKD},
  bibsource    = {dblp computer science bibliography, https://dblp.org}
}

@article{ge2024hierattr,
  author       = {Xuyang Ge and
                  Fukang Zhu and
                  Wentao Shu and
                  Junxuan Wang and
                  Zhengfu He and
                  Xipeng Qiu},
  title        = {Automatically Identifying Local and Global Circuits with Linear Computation
                  Graphs},
  journal      = {CoRR},
  volume       = {abs/2405.13868},
  year         = {2024},
  url          = {https://doi.org/10.48550/arXiv.2405.13868},
  doi          = {10.48550/ARXIV.2405.13868},
  eprinttype    = {arXiv},
  eprint       = {2405.13868},
  bibsource    = {dblp computer science bibliography, https://dblp.org}
}

@article{chalnev2024improving,
  author       = {Sviatoslav Chalnev and
                  Matthew Siu and
                  Arthur Conmy},
  title        = {Improving Steering Vectors by Targeting Sparse Autoencoder Features},
  journal      = {CoRR},
  volume       = {abs/2411.02193},
  year         = {2024},
  url          = {https://doi.org/10.48550/arXiv.2411.02193},
  doi          = {10.48550/ARXIV.2411.02193},
  eprinttype    = {arXiv},
  eprint       = {2411.02193},
  bibsource    = {dblp computer science bibliography, https://dblp.org}
}

@inproceedings{joshua2025linear,
  author       = {Joshua Engels and
                  Eric J. Michaud and
                  Isaac Liao and
                  Wes Gurnee and
                  Max Tegmark},
  title        = {Not All Language Model Features Are One-Dimensionally Linear},
  booktitle    = {The Thirteenth International Conference on Learning Representations,
                  {ICLR} 2025, Singapore, April 24-28, 2025},
  publisher    = {OpenReview.net},
  year         = {2025},
  url          = {https://openreview.net/forum?id=d63a4AM4hb},
  bibsource    = {dblp computer science bibliography, https://dblp.org}
}

@inproceedings{
farrell2024applying,
title={Applying Sparse Autoencoders to Unlearn Knowledge in Language Models},
author={Eoin Farrell and Yeu-Tong Lau and Arthur Conmy},
booktitle={Neurips Safe Generative AI Workshop 2024},
year={2024},
url={https://openreview.net/forum?id=i4z0HrBiIA}
}

@article{monroe2024mastering,
  author       = {Daniel Monroe and
                  The Leela Chess Zero Team},
  title        = {Mastering Chess with a Transformer Model},
  journal      = {CoRR},
  volume       = {abs/2409.12272},
  year         = {2024},
  url          = {https://doi.org/10.48550/arXiv.2409.12272},
  doi          = {10.48550/ARXIV.2409.12272},
  eprinttype    = {arXiv},
  eprint       = {2409.12272},
  bibsource    = {dblp computer science bibliography, https://dblp.org}
}

@article{mcgrath2023hydra,
  author       = {Thomas McGrath and
                  Matthew Rahtz and
                  J{\'{a}}nos Kram{\'{a}}r and
                  Vladimir Mikulik and
                  Shane Legg},
  title        = {The Hydra Effect: Emergent Self-repair in Language Model Computations},
  journal      = {CoRR},
  volume       = {abs/2307.15771},
  year         = {2023},
  url          = {https://doi.org/10.48550/arXiv.2307.15771},
  doi          = {10.48550/ARXIV.2307.15771},
  eprinttype    = {arXiv},
  eprint       = {2307.15771},
  bibsource    = {dblp computer science bibliography, https://dblp.org}
}

@inproceedings{marks2024sparse,
  author       = {Samuel Marks and
                  Can Rager and
                  Eric J. Michaud and
                  Yonatan Belinkov and
                  David Bau and
                  Aaron Mueller},
  title        = {Sparse Feature Circuits: Discovering and Editing Interpretable Causal
                  Graphs in Language Models},
  booktitle    = {The Thirteenth International Conference on Learning Representations,
                  {ICLR} 2025, Singapore, April 24-28, 2025},
  publisher    = {OpenReview.net},
  year         = {2025},
  url          = {https://openreview.net/forum?id=I4e82CIDxv},
  bibsource    = {dblp computer science bibliography, https://dblp.org}
}

@incollection{pearl2022direct,
  author       = {Judea Pearl},
  editor       = {Hector Geffner and
                  Rina Dechter and
                  Joseph Y. Halpern},
  title        = {Direct and Indirect Effects},
  booktitle    = {Probabilistic and Causal Inference: The Works of Judea Pearl},
  series       = {{ACM} Books},
  volume       = {36},
  pages        = {373--392},
  publisher    = {{ACM}},
  year         = {2022},
  url          = {https://doi.org/10.1145/3501714.3501736},
  doi          = {10.1145/3501714.3501736},
  bibsource    = {dblp computer science bibliography, https://dblp.org}
}

@article{dunefsky2024transcoder,
  author       = {Jacob Dunefsky and
                  Philippe Chlenski and
                  Neel Nanda},
  title        = {Transcoders Find Interpretable {LLM} Feature Circuits},
  journal      = {CoRR},
  volume       = {abs/2406.11944},
  year         = {2024},
  url          = {https://doi.org/10.48550/arXiv.2406.11944},
  doi          = {10.48550/ARXIV.2406.11944},
  eprinttype    = {arXiv},
  eprint       = {2406.11944},
  bibsource    = {dblp computer science bibliography, https://dblp.org}
}

@inproceedings{chanin2025absorption,
title={A is for Absorption: Studying Feature Splitting and Absorption in Sparse Autoencoders},
author={David Chanin and James Wilken-Smith and Tom{\'a}{\v{s}} Dulka and Hardik Bhatnagar and Satvik Golechha and Joseph Isaac Bloom},
booktitle={The Thirty-ninth Annual Conference on Neural Information Processing Systems},
year={2025},
url={https://openreview.net/forum?id=R73ybUciQF}
}

@inproceedings{
anonymous2025chessformer,
title={Chessformer: A Unified Architecture for Chess Modeling},
author={Anonymous},
booktitle={Submitted to The Fourteenth International Conference on Learning Representations},
year={2025},
url={https://openreview.net/forum?id=2ltBRzEHyd},
note={under review}
}

@article{ameisen2025circuit,
  author={Ameisen, Emmanuel and Lindsey, Jack and Pearce, Adam and Gurnee, Wes and Turner, Nicholas L. and Chen, Brian and Citro, Craig and Abrahams, David and Carter, Shan and Hosmer, Basil and Marcus, Jonathan and Sklar, Michael and Templeton, Adly and Bricken, Trenton and McDougall, Callum and Cunningham, Hoagy and Henighan, Thomas and Jermyn, Adam and Jones, Andy and Persic, Andrew and Qi, Zhenyi and Ben Thompson, T. and Zimmerman, Sam and Rivoire, Kelley and Conerly, Thomas and Olah, Chris and Batson, Joshua},
  title={Circuit Tracing: Revealing Computational Graphs in Language Models},
  journal={Transformer Circuits Thread},
  year={2025},
  url={https://transformer-circuits.pub/2025/attribution-graphs/methods.html}
}

@article{poupart2024contrastive,
  author       = {Yoann Poupart},
  title        = {Contrastive Sparse Autoencoders for Interpreting Planning of Chess-Playing Agents},
  journal      = {CoRR},
  volume       = {abs/2406.04028},
  year         = {2024},
  url          = {https://doi.org/10.48550/arXiv.2406.04028},
  doi          = {10.48550/ARXIV.2406.04028},
  eprinttype   = {arXiv},
  eprint       = {2406.04028},
  bibsource    = {dblp computer science bibliography, https://dblp.org}
}

@article{
shin2023superhuman,
author = {Minkyu Shin  and Jin Kim  and Bas van Opheusden  and Thomas L. Griffiths },
title = {Superhuman artificial intelligence can improve human decision-making by increasing novelty},
journal = {Proceedings of the National Academy of Sciences},
volume = {120},
number = {12},
pages = {e2214840120},
year = {2023},
doi = {10.1073/pnas.2214840120},
URL = {https://www.pnas.org/doi/abs/10.1073/pnas.221484012}}

@article{lindsey2024crosscoder,
  author={Jack Lindsey and Adly Templeton and Jonathan Marcus and Thomas Conerly and Joshua Batson and Christopher Olah},
  title={Sparse Crosscoders for Cross-Layer Features and Model Diffing},
  journal={Transformer Circuits Thread},
  year={2024},
  url={https://transformer-circuits.pub/2024/crosscoders/index.html}
}
\bibliographystyle{icml2026}

\newpage
\appendix
\onecolumn

\section{LC0 BT4 Details}
\label{appendix:LC0_BT4_details}

\subsection{Transformer Architecture}
\label{appendix:LC0_BT4_transformer_architecture}

LC0 is a family of transformer-based chess models trained with the Monte Carlo Tree Search (MCTS) self-play reinforcement learning paradigm introduced by AlphaZero~\cite{silver2017mastering}. 

In this work, we study BT4\footnote{Specifically, we use \texttt{BT4-1024x15x32h-swa-6147500}, a stable release from the LC0 project. The weights are publicly available at \url{https://lczero.org/play/networks/bestnets/}.}, a transformer chess model that achieves grandmaster-level performance using only its policy network~\cite{lczero2024transformer}. Our analysis focuses on this policy network; additional implementation details are shown in Figure~\ref{fig:BT4_architecture}.

BT4 consists of 15 transformer layers with residual connections and hidden size $d_{\text{model}} = 1024$. Each multi-head self-attention (MHSA) layer contains 32 attention heads. The input sequence has length 64, corresponding to the 64 squares of the $8\times8$ chessboard.

\begin{figure}
    \centering
    \includegraphics[width=0.8\linewidth]{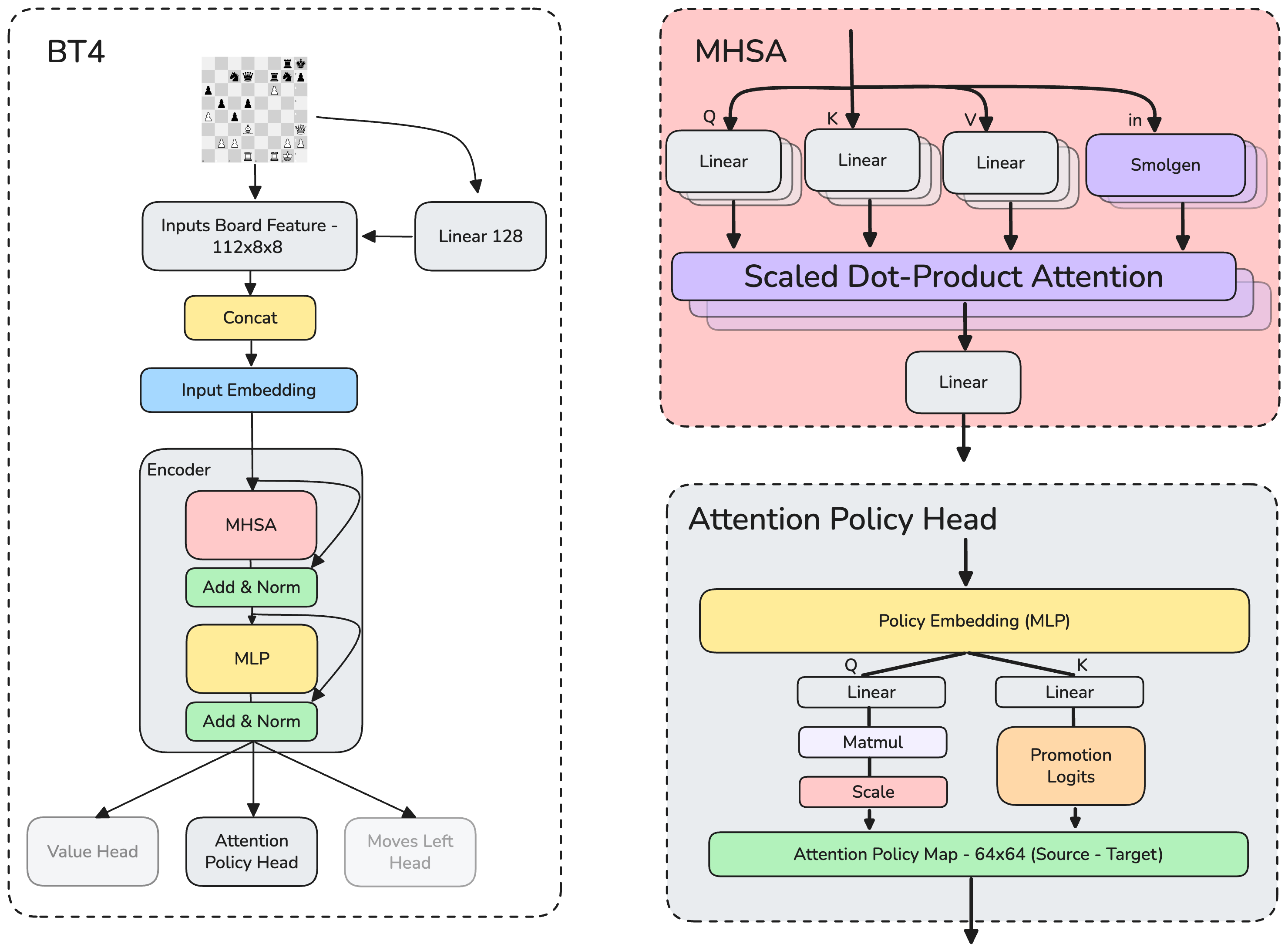}
    \caption{Overview of the BT4 architecture, featuring SmolGen-enhanced MHSA for modeling chessboard positional relations and an attention-based policy head.}
    \label{fig:BT4_architecture}
\end{figure}

\subsection{BT4 Strength Evaluation}
\label{appendix:BT4_strength}

To evaluate the playing strength of BT4 without search, we test it on Lichess puzzles of different tactical depths. Specifically, we construct five subsets corresponding to mate-in-1 through mate-in-5 puzzles. For each subset, we sample 20,000 positions, except for mate-in-5, where only 5,000 positions are used due to the smaller number of available puzzles. The results are shown in Table~\ref{tab:BT4_puzzle_accuracy}. In addition, using only a single forward pass without search of the policy network, BT4 achieved a 5--0--0 record against our 2100+ human player.

\begin{table}[h]
\centering
\small
\begin{tabular*}{\linewidth}{@{\extracolsep{\fill}}lcccccc}
\toprule
Puzzle set & Mate-in-1 & Mate-in-2 & Mate-in-3 & Mate-in-4 & Mate-in-5 & All \\
\midrule
Accuracy   & 0.996     & 0.970     & 0.890     & 0.776     & 0.707     & 0.932 \\
\bottomrule
\end{tabular*}
\caption{BT4 puzzle accuracy without search on Lichess mate puzzles of varying tactical depth. We sample 20,000 positions for each subset from mate-in-1 to mate-in-4, and 5,000 positions for mate-in-5 due to limited data availability.}
\label{tab:BT4_puzzle_accuracy}
\end{table}

\section{Sparse Decomposition Details}
\label{appendix:sparse_decomposition_details}
We configure the Transcoder and Lorsa modules with $k=30$ and an expansion factor of 16. This results in $1024 \times 16 = 16{,}384$ Transcoder and Lorsa features per layer, with 30 activated per square. Lorsa and Transcoder features are both intrinsically 1-dimensional and can be represented uniformly via encoder and decoder vectors, so we match the number of them. To decrease the number of dead features, we apply an auxiliary $k$ loss~\cite{gao2024oaisae} for both Transcoders and Lorsas.

For Lorsas, in order to preserve fidelity to the original MHSA computation, we match the attention architecture described in Appendix~\ref{appendix:LC0_BT4_transformer_architecture}, including bilinear attention and the Smolgen module. We further optimize the initialization scheme by initializing $W_O$ within the output subspace of the original attention heads~\cite{wang2025attention}. The original $W_Q$, $W_K$, Smolgen parameters, and attention scaling factors are also used to initialize Lorsas.

\section{Faithfulness of Sparse Replacement Layers}
In Figure~\ref{fig:TC_Lorsa_MSE}, we examine the \textbf{L2 Norm Error Ratio} and \textbf{Explained Variance(EV)} of the Transcoders and Lorsas we trained in random sampled dataset.

\begin{figure}[H]
    \centering
    \begin{subfigure}[t]{0.49\linewidth}
        \centering
        \includegraphics[width=\linewidth]{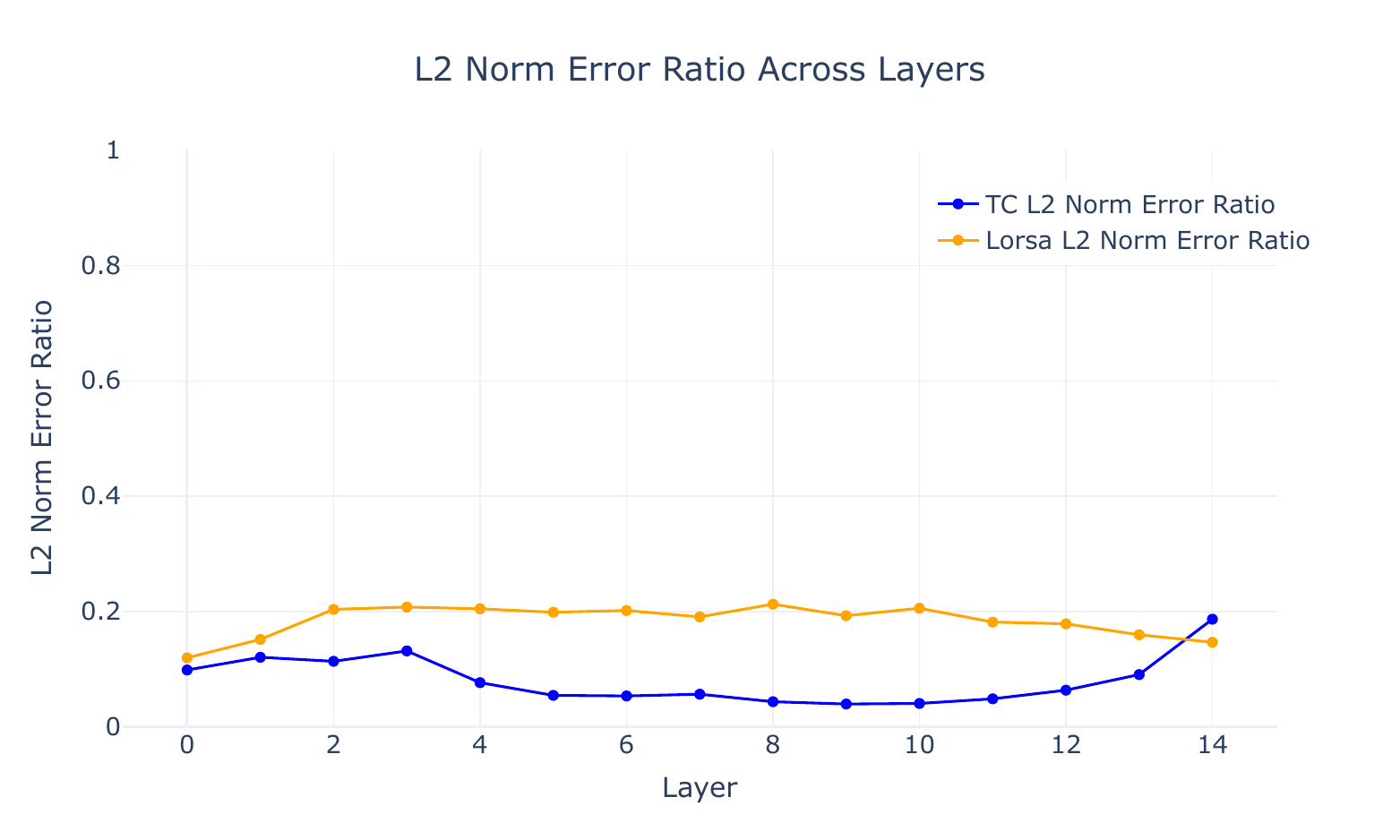}
        \caption{Layer-wise L2 norm error ratio of sparse replacement layers.}
        \label{fig:l2_error_across_layers}
    \end{subfigure}
    \hfill
    \begin{subfigure}[t]{0.49\linewidth}
        \centering
        \includegraphics[width=\linewidth]{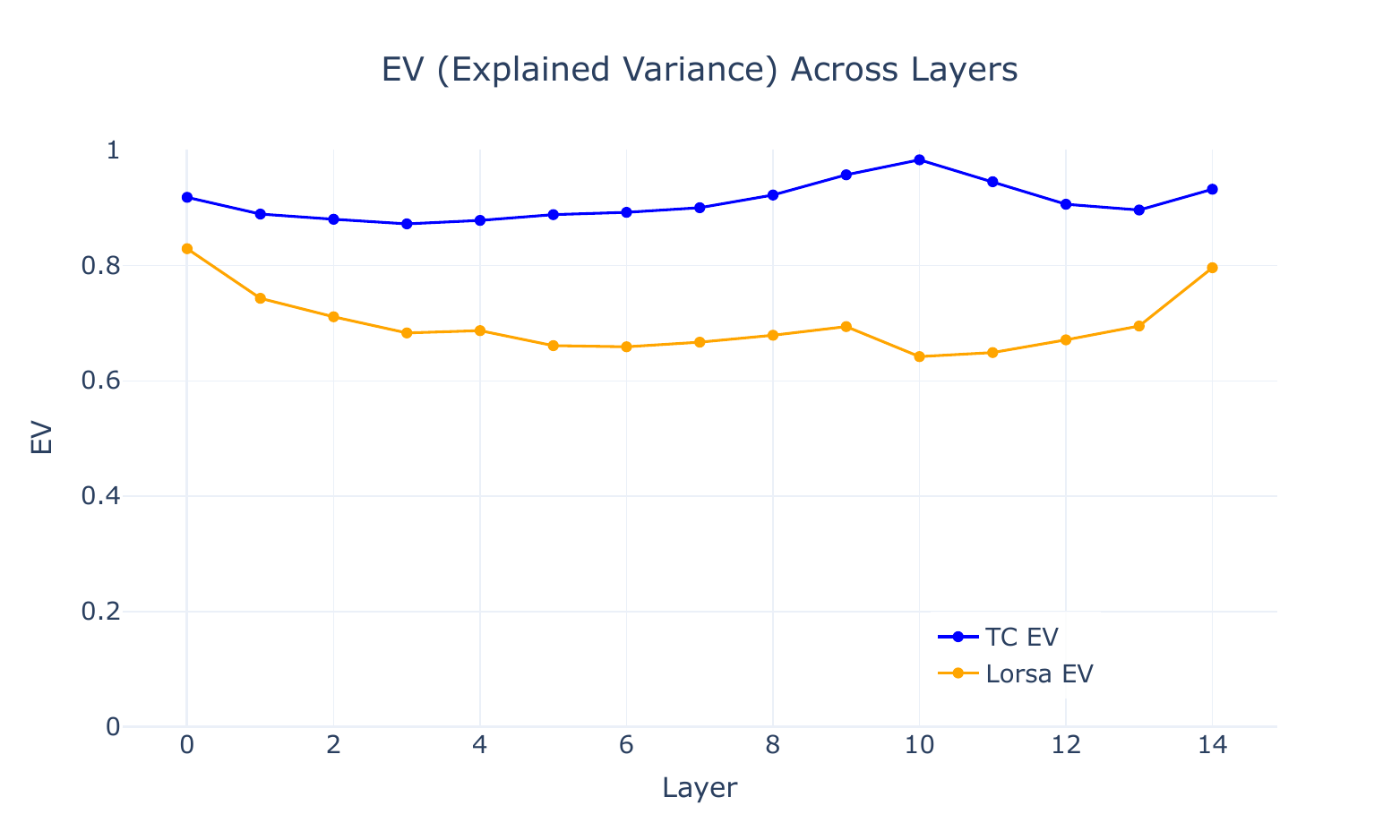}
        \caption{Layer-wise Explained variance (EV) of sparse replacement layers.}
        \label{fig:ev_across_layers}
    \end{subfigure}
    \caption{
    Faithfulness of sparse replacement layers.
    We report the L2 norm error ratio (left) and the explained variance EV (right) of Transcoder and Lorsa modules across transformer layers on a randomly sampled dataset.
    }
    \label{fig:TC_Lorsa_MSE}
\end{figure}

\begin{align}
\text{L2 Norm Error Ratio} = \frac{\mathbb{E}_t\left[\|\hat{\mathbf{x}}_t - \mathbf{x}_t\|_2\right]}{\mathbb{E}_t\left[\|\mathbf{x}_t\|_2\right]}.
\end{align}

\begin{align}
\text{Explained Variance} = 1 - \frac{\mathbb{E}_t\left[\|\hat{\mathbf{x}}_t - \mathbf{x}_t\|_2^2\right]}{\mathbb{E}_t\left[\|\mathbf{x}_t - \bar{\mathbf{x}}\|_2^2\right]}.
\end{align}

As shown in ~\ref{fig:TC_Lorsa_MSE}, these metrics quantify how accurately sparse replacement layers reconstruct the original module output. We observe that all modules have an L2 norm error ratio below 0.3 and explained variance above 0.6. While Lorsas show lower explained variance in the middle layers—particularly for attention outputs—this is consistent with the increased architectural complexity and nonlinearity of information transfer at intermediate depths.

\section{Formalism of the Interpretation Language}
\label{appendix:interpretation_explain}
Table~\ref{tab:feature_validation} uses a compact, semi-formal rule language to describe the manual verification criteria for interpretable Transcoder and Lorsa features.
We clarify the notation and conventions below.

\paragraph{Entity Shorthands.}
We use abbreviated prefixes to denote chess entities: Opp. refers to the opponent's pieces (e.g., Opp.Knight, Opp.Queen, Opp.Pieces), while Own refers to the current player's pieces.
\paragraph{Activation Predicate.}
The predicate \texttt{Act.@X} indicates that the feature is activated at squares associated with entity $X$. For example, \texttt{Act.@Opp.Knight} denotes that the feature is activated at squares of an opponent's knight, while \texttt{Act.@Pawn x Opp.Piece} denotes activation on capture target squares where a pawn captures an opponent piece.

\paragraph{Directional Attention Operator.}
The arrow notation $\leftarrow$ indicates a $z$-pattern direction for a Lorsa feature. Specifically, \texttt{A $\leftarrow$ B} means that the feature activation at location $A$ attends to, or is largely contributed by, entity $B$. For example, \texttt{Act. @ Opp.Queen $\leftarrow$ OwnQueen} indicates that the feature activation at the opponent's queen is contributed through a strong $z$-pattern by the current player's own queen.

\paragraph{Spatial Relations.}
Spatial relations are expressed using intuitive predicates such as:
\texttt{Adj.at X} for squares adjacent to entity $X$, \texttt{Surrounding} for the eight neighboring squares around a piece, and directional descriptors indicate orientation on the board: \texttt{rank-wise} refers to squares along the same rank (row) as the entity, while \texttt{diagonal} refers to squares along the diagonal lines through the entity.

\paragraph{Complex Piece Dependencies.}
Certain features capture more intricate piece relationships, which are expressed as additional constructs in the interpretation language. The predicate \texttt{n-move reachability} indicates that a piece can legally reach the activated square within $n$ moves; for example, \texttt{2-move reachability} denotes activation on squares reachable by a piece within two moves. The predicate $\le 2$ \texttt{blockers} specifies that there are at most two intervening pieces between a source piece and the target square. The \texttt{xchg} predicate denotes that a piece is in an exchange state, such as being part of a potential capture or tactical trade. These constructs allow the interpretation language to capture relational and tactical dependencies beyond simple adjacency or direct attacks, while remaining human-interpretable and consistent with the activation-based feature analysis framework.

\paragraph{Composite Conditions.}
Multiple conditions within a rule are interpreted conjunctively.
For example, a rule combining adjacency, protection, and check status indicates that all listed conditions must be satisfied for a square to be counted as a positive match.
These composite rules are designed to remain human-interpretable while capturing non-trivial tactical patterns.

\paragraph{Design Rationale.}
The interpretation language and validating rules are intentionally concise rather than fully formal. They are designed to provide clear and reproducible criteria for semantic validation, while remaining human-readable.

Due to the rich and highly compositional nature of chess, and the potential mismatch between a model’s internal representations and human-intuitive, textbook-style board annotations, some visual and relational patterns inevitably require high-level abstractions. As shown in Table~\ref{tab:feature_validation}, several features achieve high precision but relatively lower recall, which we attribute to the coarse granularity of the current rule descriptions.

Designing precise interpretation rules that reflect the true distribution of activations remains a challenging problem. Balancing abstraction and coverage in these rules is part of ongoing work in mechanistic interpretability in chess.

\section{Representative Feature Examples by Category}
\label{appendix:taxonomy_feature}

Figure~\ref{fig:categories} illustrates representative examples of features across different categories, as defined in our interpretation language. We describe each category below, corresponding to subfigures (1)–(7).

\paragraph{Detection Features [Det]:}
Detection features are designed to identify the presence of a specific kind of piece on the board. For example, in Figure~\ref{fig:categories}(1), \texttt{Tc.0.5593} detects an opponent's knight. These features typically activate on squares of a specific piece type.

\paragraph{Source and Target Square Features [Src / Tgt]:}
Features that activate on the \emph{source square} of a predicted move are labeled as [Src], while those on the \emph{target square} are labeled as [Tgt]. These features mostly appear in later layers. Figure~\ref{fig:categories}(2) shows an example of a [Src] feature.

\paragraph{Value-Based Features [Val]:}
Value-based features activate on squares with a specific positional value. In Figure~\ref{fig:categories}(3), a feature is activated at squares of ``high-value chessboards''.

\paragraph{Capture Features [Cap]:}
Capture features identify squares where a piece can capture an opponent's piece, typically a rook, queen, or pawn. Figure~\ref{fig:categories}(4) illustrates a [Cap] feature representing a potential capture of an opponent's rook.

\paragraph{Tactical Features [Tac]:}
Tactical features capture more complex tactics on the chessboard. For example, in Figure~\ref{fig:categories}(5), a [Tac] feature represents a scenario in which our rook is lost due to a check on the king. These features encode interactions that are more intricate than simple captures.

\paragraph{Spatial Relation Features [Spa]:}
Spatial relation features describe activation patterns that are related to fixed regions on the board, the area surrounding a piece, or the coverage area of a piece. In Figure~\ref{fig:categories}(6), a [Spa] feature is activated at all squares of the same color as the king, capturing spatial patterns across the board.

\paragraph{Piece Movement Features [Mov]:}
Piece movement features describe the movement or control patterns of pieces. In Figure~\ref{fig:categories}(7), a [Mov] feature is activated at the eight squares surrounding the king, representing either the king's potential movement area or its coverage influence.

\begin{figure}
    \centering
    \includegraphics[width=0.9\linewidth]{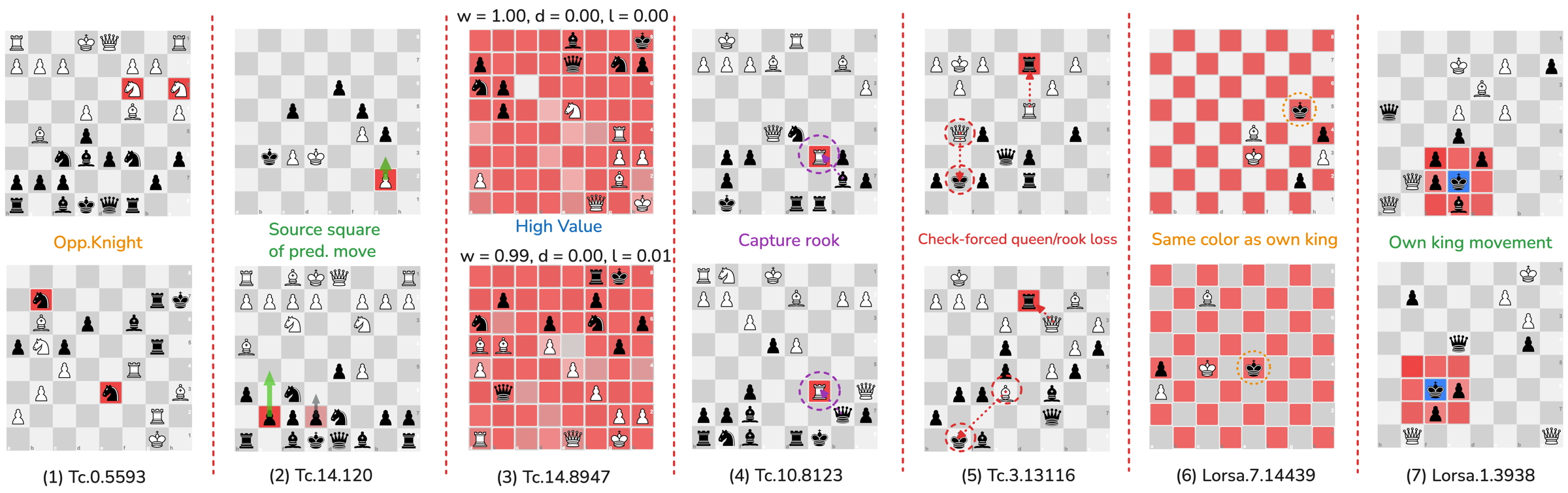}
    \caption{Representative feature examples by category: 
    (1) Detection feature [Det] activating on squares occupied by an opponent knight; 
    (2) Source square feature [Src] activating on the square from which the predicted move originates; 
    (3) Value-based feature [Val] activating on high-value positions on the board; 
    (4) Capture feature [Cap] activating on squares from which a piece can capture an opponent's rook; 
    (5) Tactical feature [Tac] activating in scenarios where our rook is lost due to a check; 
    (6) Spatial-related feature [Spa] activating on squares reflecting board-wide spatial patterns, e.g., squares of the same color as the king; 
    (7) Piece movement feature [Mov] activating on squares representing the movement or coverage area of a piece, here the eight squares surrounding the king.}
    \label{fig:categories}
\end{figure}

\section{Algorithm Details}
\label{appendix:reasoning_path_generation}

\begin{figure}[H]
    \centering
    \begin{subfigure}[t]{0.49\linewidth}
        \centering
        \includegraphics[width=\linewidth]{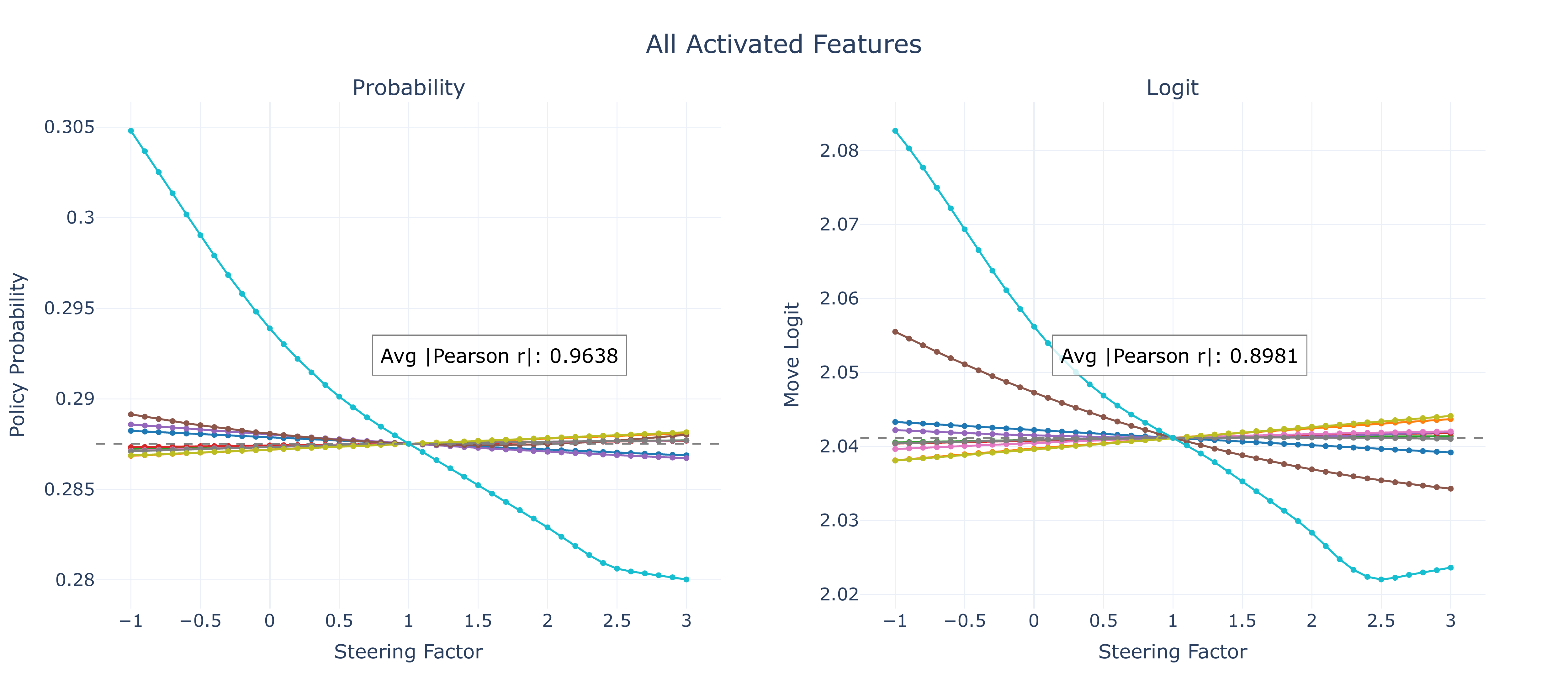}
        \caption{Randomly ten features sampled from all activated features from a single forward pass generally have a small effect on policy probability but exhibit a strong linear relationship with the steering factor.}
        \label{fig:steering_activated_features}
    \end{subfigure}
    \hfill
    \begin{subfigure}[t]{0.49\linewidth}
        \centering
        \includegraphics[width=\linewidth]{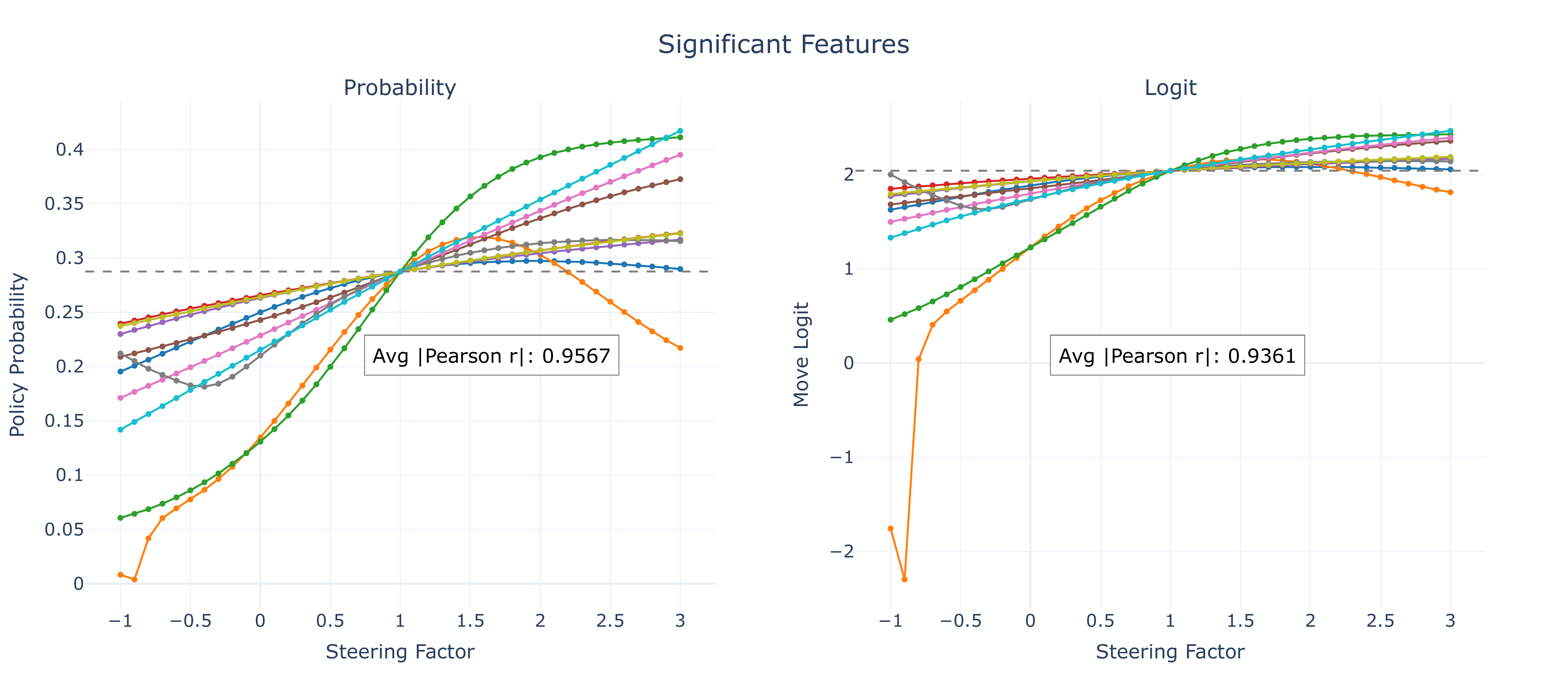}
        \caption{Randomly ten features sampled from the top 100 significant features under steering factor $\alpha = -1$ still show a high linear relationship with policy output, with an average absolute Pearson correlation of 0.9567, even higher linear correlation in the range $[-1, 0]$.}
        \label{fig:steering_significant_features}
    \end{subfigure}

    \caption{Effect on policy probability and logit of steering factor on feature activations. (a) Ten random activated features at each steering factor and corresponding change of policy probability and logit. (b) Ten random significant features at each steering factor and corresponding change of policy probability and logit.}
    \label{fig:steering_scale_features}
\end{figure}

We mainly focus on moves with probabilities greater than $0.25$, as they are less susceptible to noisy interference, enabling clearer visualization and analysis of the reasoning pathways.
\subsection{Steering Factor Analysis.}
\label{appendix:steering_factor}
In Algorithm~\ref{alg:reasoning_path_generation}, we apply feature steering with factors $\alpha = -1$ and $\beta = -1$, corresponding to zeroing out the feature's contribution in the residual stream. For a reasoning path in the case study described in Section~\ref{sec:case_study}, we randomly sample 10 features from both the significant feature set and all activated features. We then plot how the probability of move \textcolor{BestMove}{\textbf{\texttt{Qxh7+}}} changes as a function of the steering factor, as shown in Figure~\ref{fig:steering_scale_features}. The mean absolute Pearson correlation for all samples is 0.9985, indicating a strong linear relationship between feature activation and policy output within the tested range. Based on this observation, we select a steering factor of $\alpha = -1$ and randomly choose 10 features from the top 100 significant features. These features also exhibit a high degree of linearity, with an average absolute Pearson correlation of 0.9567, particularly within the steering factor range $[-1, 0]$. This demonstrates that our choice of $\alpha = -1$ lies well within the linear regime, making it a reasonable setting for causal feature intervention.

To ensure that the selection of significant features is not overly sensitive to the choice of steering factor, we evaluate the Pearson correlation between steering factors and feature–output effects. 

Specifically, we randomly sample 100 positions from the dataset, consider the top-1 policy probability for each position, and randomly select 10 features per sample. Steering factors are varied over the range $[-2, 2]$. We observe a high absolute Pearson correlation of $0.96 \pm 0.13$ (full distribution shown in Figure~\ref{fig:pearson_correlation_distribution}), indicating that, within this range, feature effects on the policy output are approximately linear. And the number of features that promote and suppress the output probability are comparable. Consequently, our choice of $\alpha = -1$ is intuitive and empirically justified, as it falls within the linear regime where feature contributions scale proportionally with the steering factor.

\begin{figure}[H]
    \centering
    \includegraphics[width=0.6\linewidth]{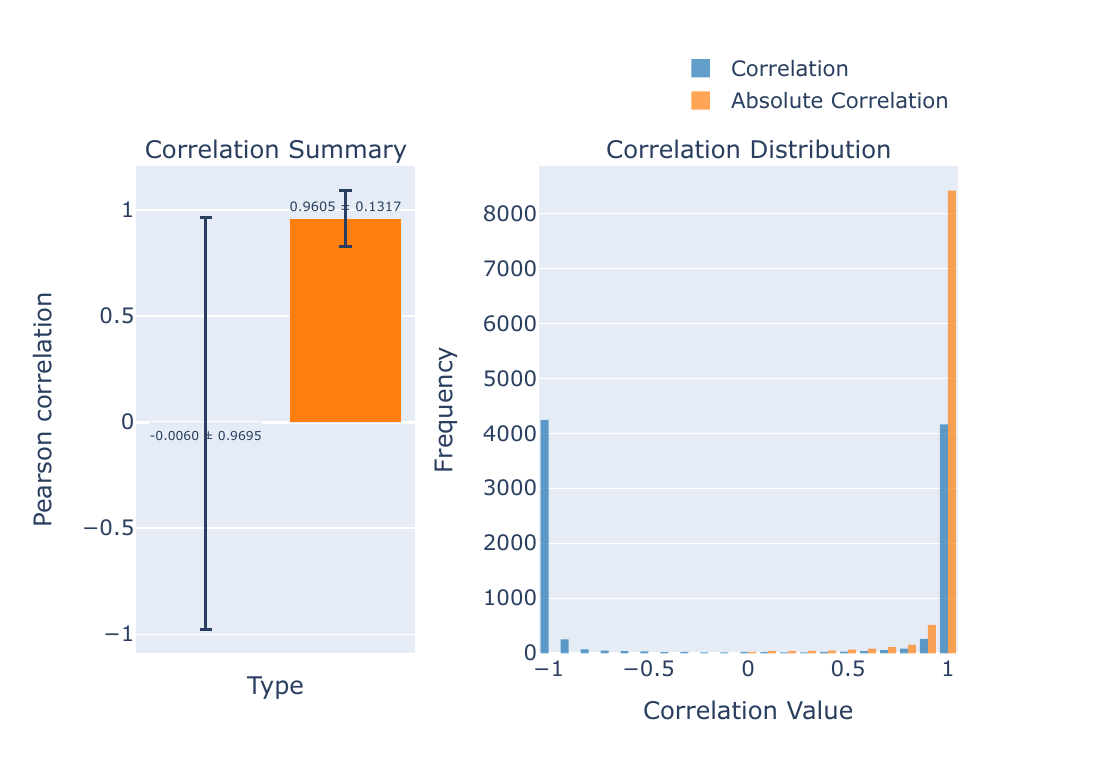}
    \caption{
    Pearson correlation distribution of steering sensitivity.
    (Left) The impact of randomly selected features on the top policy probability within the $[-2, 2]$ steering range.
    (Right) The distribution of Pearson correlation coefficients.
    We observe significant clustering in the ranges $[-1, -0.9]$ and $[0.9, 1]$, with comparable counts.
    }
    \label{fig:pearson_correlation_distribution}
\end{figure}

\subsection{Feature and Edge Pruning.}
\label{appendix:pruning}
This amplifies the observable impact of individual features on policy probabilities while reducing interference. For \emph{feature-feature effect}~\ref{eq:feature_effect} analysis, we directly remove the feature contribution from the residual stream, allowing us to capture genuine local repair behaviors (i.e., the hydra effect).

For feature pruning, in the case study (Section~\ref{sec:case_study}), we retain the top 200 features
with the largest influence on a given move.
For edge pruning, we keep feature--feature effects satisfying
$\mathrm{Effect}(f_i \!\rightarrow\! f_j) > 0.1 \cdot a_j^{(h^l)}$,
which accounts for variation in feature activation magnitudes and yields a balanced sparsification.
This approach aligns with similar methodologies established in \cite{ameisen2025circuit, voita2019analyzing}, ensuring that our selection criteria are principled rather than an instance of cherry-picking.

\section{Universality Evaluation Dataset}
\label{appendix:universal_datasets}
To verify that path-level universal characteristics are not confined to isolated examples, we conduct our analysis on filtered evaluation datasets with varying levels of uncertainty in the policy output.

Let $p_{(1)}$ and $p_{(2)}$ denote the model-predicted probabilities of the top-1 and top-2 legal moves, respectively.
We define the policy margin as
\[
\Delta p = p_{(1)} - p_{(2)}.
\]

We retain positions satisfying
\[
\mathcal{F}=\{\text{Position}\in\mathcal{D}\mid p_{\min}<\Delta p<p_{\max}\}.
\]

To assess how uncertainty impacts reasoning, we evaluate three regimes defined by the policy margin $\Delta p$:
\textbf{All} ($\Delta p \in [0, 1]$),
\textbf{Confident} ($\Delta p \in [0.8, 1]$),
and \textbf{Confused} ($\Delta p \in [0, 0.2]$).
Additionally, we introduce a specialized setting for comparison in which the top candidate moves share the same source square, referred to as \textbf{Same-source}.

\section{Supplementary Details for the Case Study}

\begin{figure}
    \centering
    \includegraphics[width=0.65\linewidth]{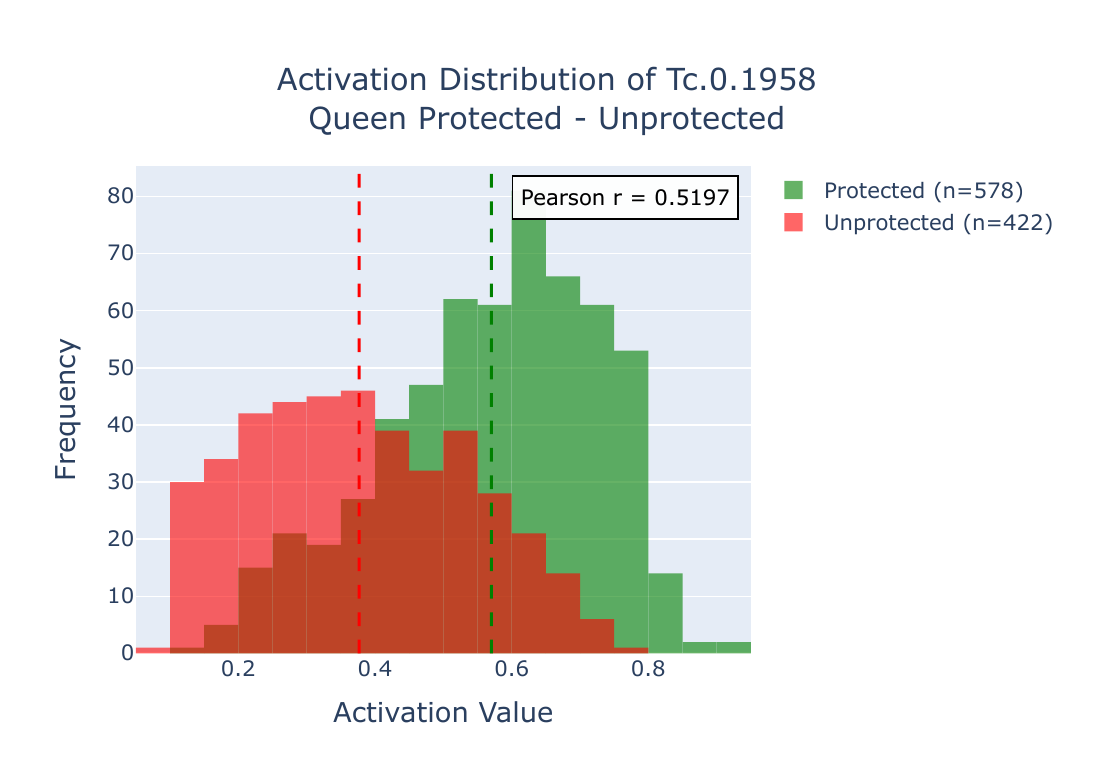}
    \caption{The correlation of between feature activation and whether the exchanging-queen is protected in 1000 activation times with the exchanging-queen. The activation value exhibits a Pearson correlation of 0.52.}
    \label{fig:tc_10_1958_protected}
\end{figure}

\subsection{The Geometry Similarity of Bishop-movement Lorsa features}
\label{appendix:cl_superposition}

As mentioned in the case study in Section~\ref{sec:case_study}, we identify three Lorsa features responsible for propagating bishop control information, \textbf{\texttt{Lorsa.1.8260}}, \textbf{\texttt{Lorsa.7.14502}} and \textbf{\texttt{Lorsa.8.9586}}. We further projected decoder vectors of these three Lorsa features onto the final residual stream given LC0’s Post-LayerNorm architecture shown in Appendix~\ref{appendix:LC0_BT4_transformer_architecture}, to reveal their shared semantic direction. As shown in Table~\ref{tab:lorsa_cosine_matrix}, the high pairwise cosine similarities confirm that these distributed Lorsa features exhibit a cross-layer superposition. The method we amplified the steering factor to suppress or enhance a particular function, rather than manipulating the entire feature group~\cite{ameisen2025circuit} is reasonable.

\begin{table}[t]
\centering
\caption{Cosine similarity matrix of decoder vectors for bishop-movement Lorsa features (after LC0 post-LayerNorm). High off-diagonal cosine similarities indicate a shared semantic direction across distributed layers.}
\label{tab:lorsa_cosine_matrix}
\small
\resizebox{0.48\columnwidth}{!}{%
\begin{tabular}{lccc}
\toprule
 & \textbf{\texttt{Lorsa.1.8260}} & \textbf{\texttt{Lorsa.7.14502}} & \textbf{\texttt{Lorsa.8.9586}} \\
\midrule
\textbf{\texttt{Lorsa.1.8260}}   & 1.00 & 0.82 & 0.84 \\
\textbf{\texttt{Lorsa.7.14502}}  & 0.82 & 1.00 & 0.97 \\
\textbf{\texttt{Lorsa.8.9586}}   & 0.84 & 0.97 & 1.00 \\
\midrule
\multicolumn{4}{c}{\textbf{Random baseline:} $\mathbb{E}[\cos \theta] = 0.17$} \\
\bottomrule
\end{tabular}
}
\end{table}

\subsection{Highest-Activation Samples for Features in Case Study}
\label{appendix:case_features}
In this subsection, we present two board position samples with the highest activations for each feature discussed in the case studies. These positions are not necessarily from the same game or tactical context, as they are independently sampled from the dataset during feature activation analysis.

\begin{figure}[H]
    \centering
    \includegraphics[width=0.65\linewidth]{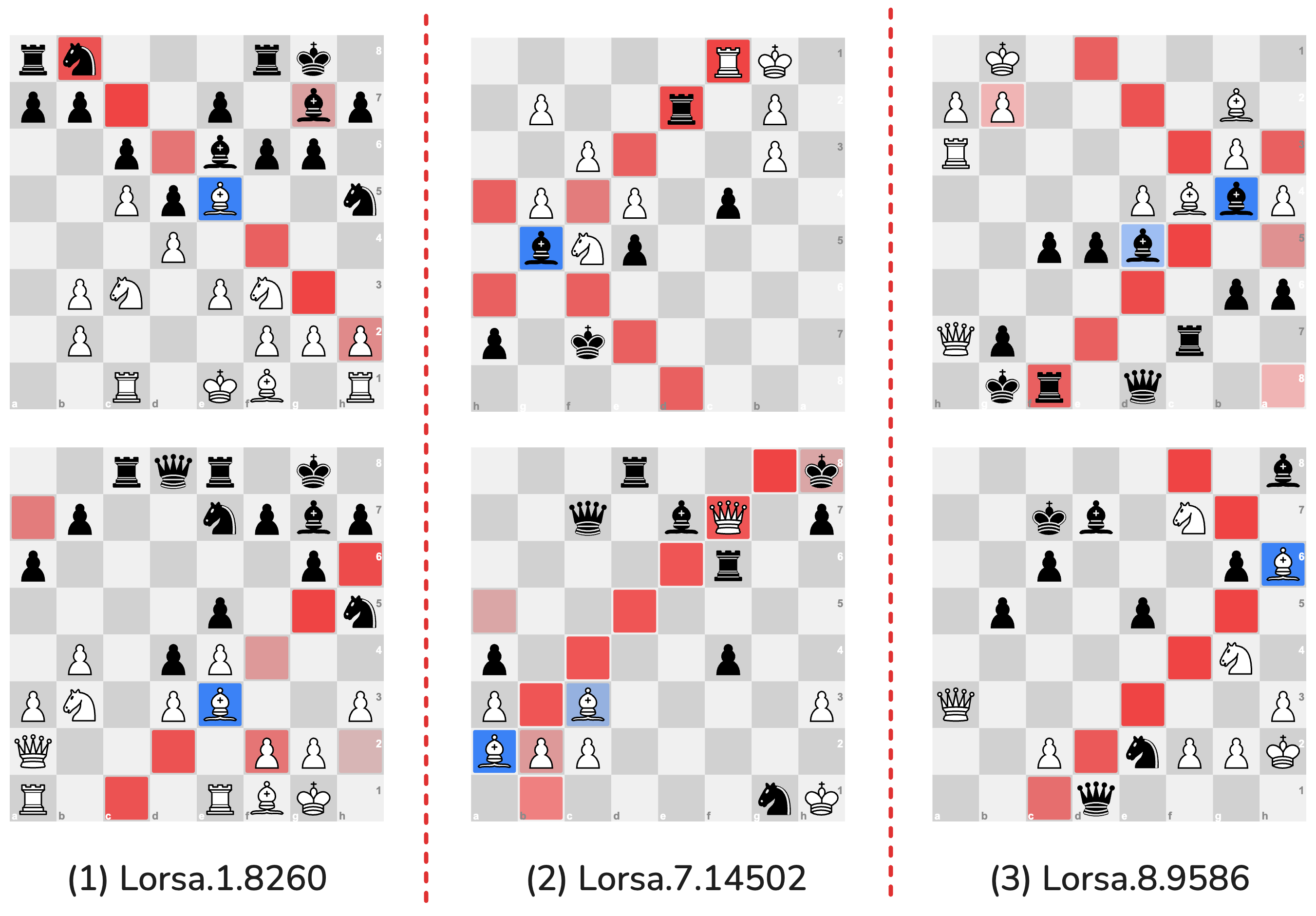}
    \caption{samples with the highest activations of three bishop-move Lorsa features mentioned in Section~\ref{sec:case_study} finding 1.}
    \label{fig:finding1_features}
\end{figure}

\begin{figure}[H]
    \centering
    \includegraphics[width=0.99\linewidth]{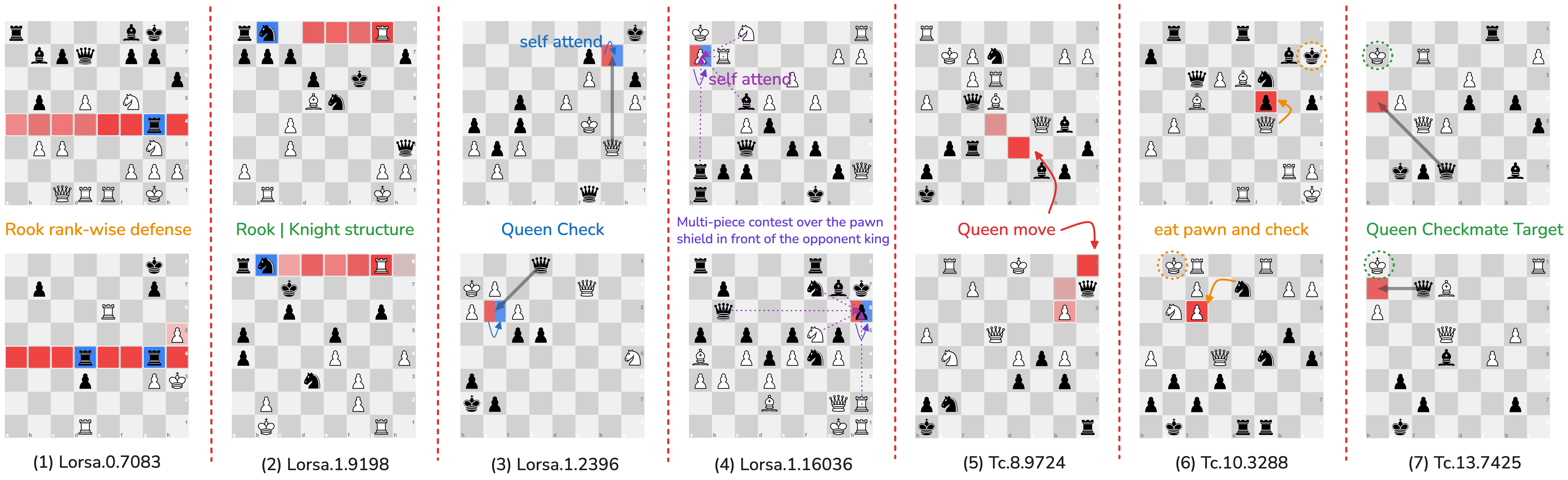}
    \caption{samples with the highest activations of features mentioned in Section~\ref{sec:case_study} finding 2.}
    \label{fig:finding2_features}
\end{figure}

For \emph{Finding 1}, the samples with the highest activations of the three bishop-movement Lorsa features are shown in Figure~\ref{fig:finding1_features}. These features exhibit similar activation and corresponding $z$-patterns, effectively transferring information about the bishop’s control to the squares under its coverage. They are distributed across layers and exhibit an important role in encoding piece control information. These feature visualizations give a solid validation to our findings and generality.

For \emph{Finding 2}, we show that activation copying at the \textbf{\texttt{h7}} square directly affects a set of downstream significant features visualized in Figure~\ref{fig:finding2_features}. These features encode semantics that are highly relevant to tactical patterns and final decision-making. Through rule-based validation, we confirm that they exhibit strong logical structure. For example, changes in the activation of \textbf{\texttt{Lorsa.1.16036}} indicate that the model’s assessment of the attack–defense balance at the \textbf{\texttt{h7}} pawn has been disrupted, reflecting a decisive shift in the local tactical evaluation.

\begin{figure}[H]
    \centering
    \includegraphics[width=0.6\linewidth]{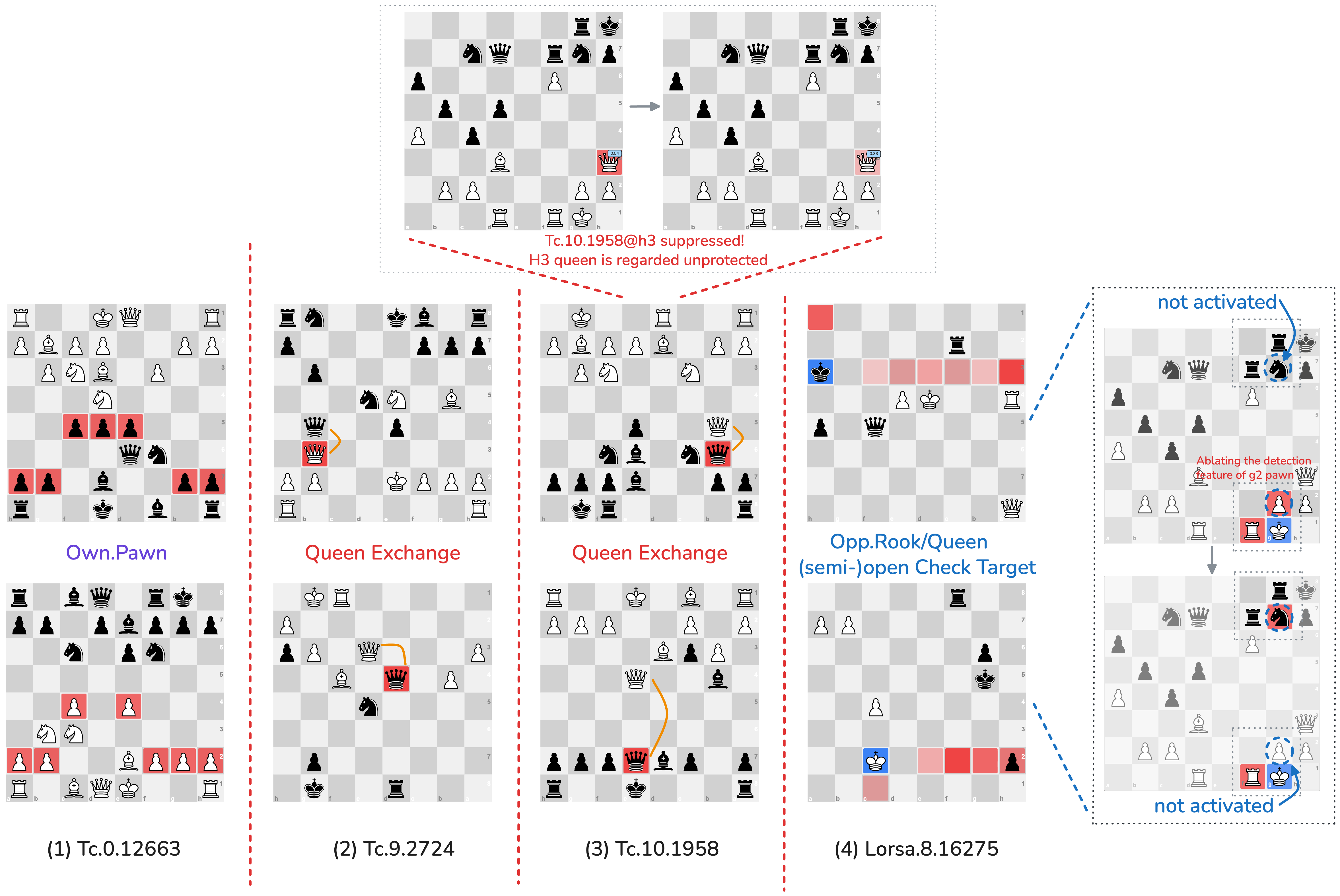}
    \caption{samples with the highest activations of features mentioned in Section~\ref{sec:case_study} finding 3.}
    \label{fig:finding3_features}
\end{figure}

\begin{figure}[H]
    \centering
    \includegraphics[width=0.65\linewidth]{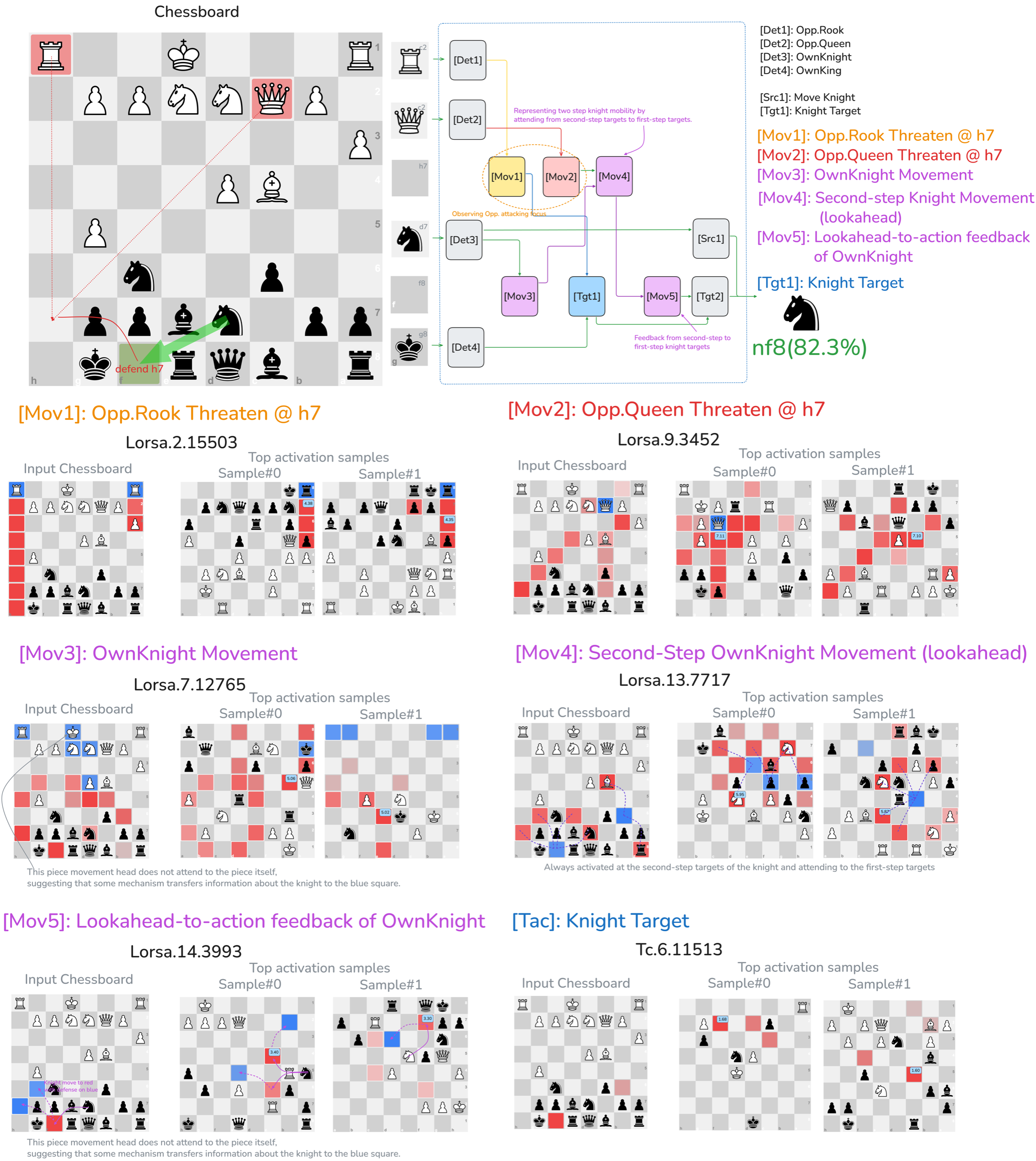}
    \caption{
    Reasoning pathway for a defensive decision.
    The model detects an early-layer threat from the opponent’s queen--rook coordination threatening at \textbf{\texttt{h7}}, and consequently mobilizes a knight to block the mate-in-one threat.
    Key features involved in this defensive reasoning are visualized.
    }
    \label{fig:defense}
\end{figure}

In \emph{Finding 3}, we observe that the removal of \textbf{\texttt{g2}} pawn significantly alters the model's policy output. This effect is closely mirrored by the ablation of Transcoder \textbf{\texttt{Tc.0.12663@g2}}, suggesting that this feature serves as a primary neural representation of the g2 pawn. And ablating this \emph{pawn-detection feature} leads to a decrease in the downstream of \emph{queen exchange} (e.g. \textbf{\texttt{Tc.10.1958@h3}}) feature.  

This feature is activated during queen exchanges, and we find that its activation is highly correlated with whether the own queen is protected under exchange situations (Figure~\ref{fig:tc_10_1958_protected}). When its activation diminishes (as seen in Figure \ref{fig:tc_10_1958_protected}), the model tends to perceive the queen on \textbf{\texttt{h3}} as unprotected, encouraging queen movement.

This activation shift from g2 to g7 suggests that, without the protection of the g2 pawn, the model anticipates a heightened vulnerability to checks on g7. Consequently, it evaluates \textcolor{Suboptimal}{\textbf{\texttt{fxg7+}}} as detrimental to king safety. 

In the presence of the g2 pawn, the model perceives its queen protected and defensive posture as stable. This internal state biases the model toward highly risk-averse strategies, a phenomenon we term "defensive fixation."

\section{More Cases of Reasoning Pathways}
\label{appendix:case_supernode}
\paragraph{Defensive Reasoning}
In this case shown in Figure~\ref{fig:defense}, the model identifies a coordinated threat between the opponent’s queen and rook that could lead to a checkmate at \textbf{\texttt{h7}}. In response, the model reroutes a knight to defend the square, with the relevant information propagated through a knight-movement feature. Specifically, the model uses \textbf{\texttt{Lorsa.13.7717}} to transmit the information that “the knight can defend \textbf{\texttt{h7}},” anchoring this defensive signal at \textbf{\texttt{h7}}. The associated lookahead information is then fed back via \textbf{\texttt{Lorsa.14.3993}}, propagating from the defended square to the action (target) square position. We identify the process of \emph{lookahead-to-action} mechanism through our reasoning pathway.

\paragraph{Offense with Concurrent Defensive Awareness.}
In this case shown in Figure~\ref{fig:win_or_go_home}, the model selects the correct policy \textcolor{BestMove}{\textbf{\texttt{Qc4+}}}.
Analysis of the corresponding reasoning pathway reveals that, while pursuing an immediate attacking continuation, the model simultaneously accounts for a potential mate-in-one threat from the opponent’s queen. The square \textbf{\texttt{c4}} not only enables a diagonal attack on the opposing king but also effectively neutralizes this imminent defensive risk.  
We visualize the local reasoning pathway supporting this decision, highlighting key features associated with opponent mating threats and diagonal attack patterns, as summarized by the relevant supernodes.

\begin{figure}[H]
    \centering
    \includegraphics[width=0.75\linewidth]{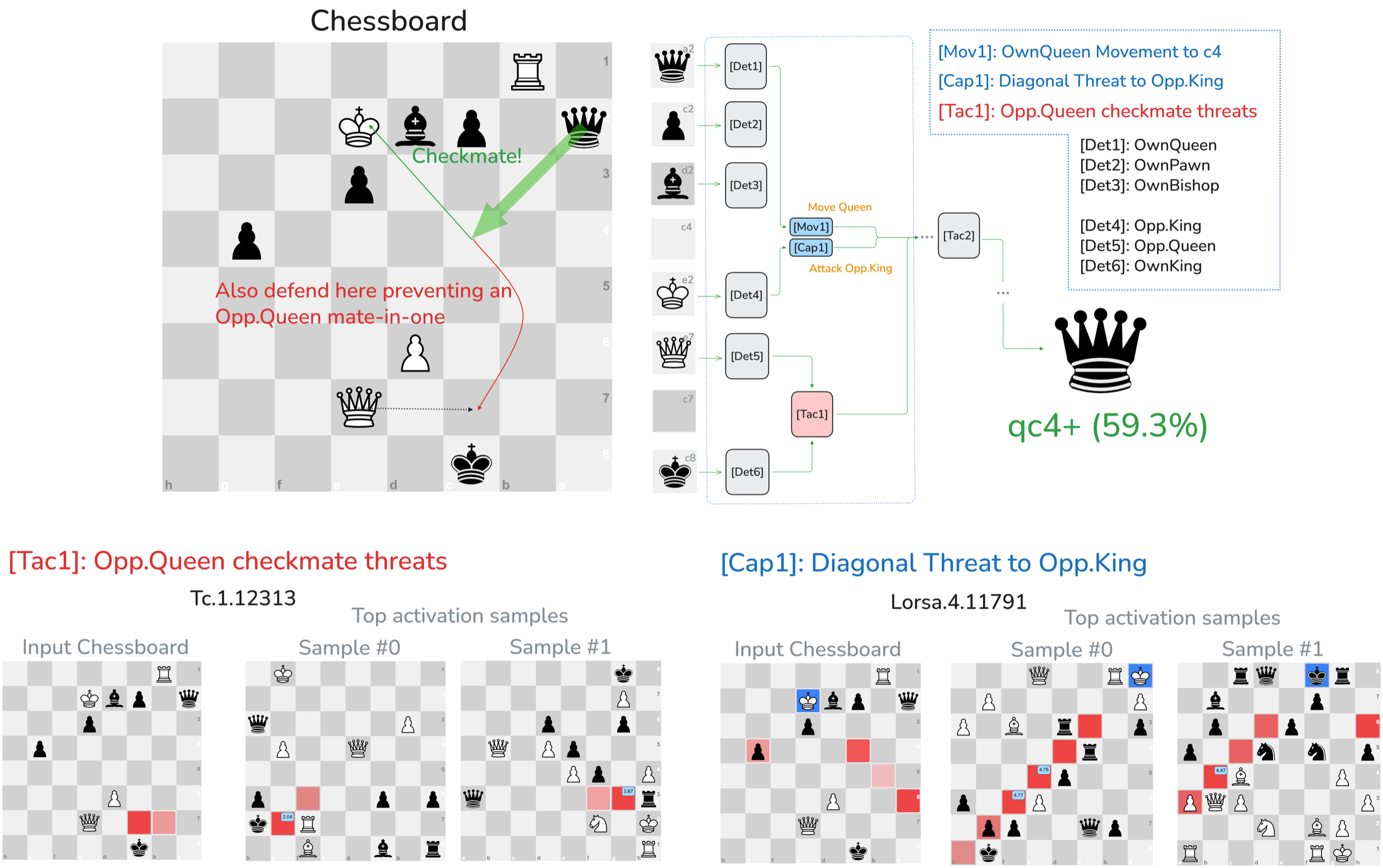}
    \caption{
    Reasoning pathway of a case where the model combines offensive and defensive reasoning.
    The figure shows the input chessboard position, the reasoning pathway supporting the selected policy \textcolor{BestMove}{\textbf{\texttt{Qc4+}}}, and the associated supernode interpretations.
    Two key features are visualized: \textbf{\texttt{Tc.1.12313}} represents the awareness of opponent queen's checkmate threats, and \textbf{\texttt{Lorsa.4.11791}} exerting diagonal threats against the opposing king.
    }
    \label{fig:win_or_go_home}
\end{figure}

\paragraph{Look-ahead Reasoning in a Mate-in-two Sequence via Tactical Sacrifice.}
\begin{figure}[H]
    \centering
    \includegraphics[width=0.75\linewidth]{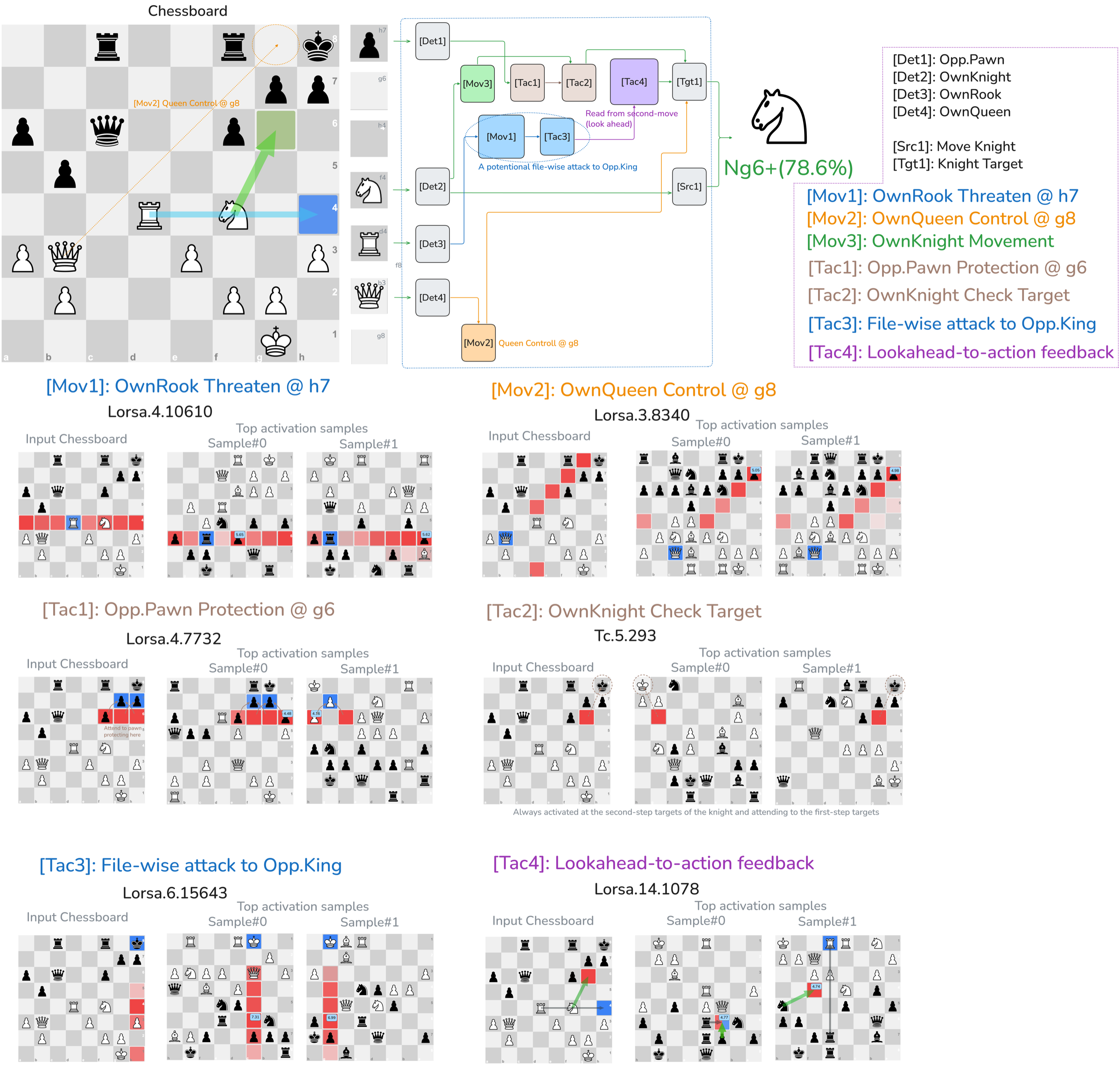}
    \caption{Reasoning pathway of look-ahead reasoning in a mate-in-two scenario. The reasoning pathway illustrates how the model integrates the immediate sacrifice move (\textbf{\texttt{Ng6+}}) with the anticipation of a subsequent checkmate threat from the rook at \textbf{\texttt{h4}}. Important correlated features with highly semantic interpretability are visualized below.}
    \label{fig:looking_ahead}
\end{figure}
In this case depicted in Figure~\ref{fig:looking_ahead}, the model correctly identifies a valid mate-in-two sequence and executes the move of sacrificeing the knight \textcolor{BestMove}{\textbf{\texttt{Ng6+}}}. The reasoning pathway of \textcolor{BestMove}{\textbf{\texttt{Ng6+}}} reveals that while the model encodes the defensive coverage of the opponent's \textbf{\texttt{h7}} pawn on the \textbf{\texttt{g6}} target square, simultaneously a cluster of rook-attack features at \textbf{\texttt{h4}} is activated, representing the subsequent checkmate move. This tactical foresight is propagated back to the \textbf{\texttt{g6}} target representation through features such as \textbf{\texttt{Lorsa.14.1078}}. This discovery corroborates the evidence of learned look-ahead reported in \cite{jenner2024evidence}, while providing a more granular mechanistic account of how such tactical foresight is computed.

\paragraph{A complex middlegame sacrifice case.}

\begin{figure}[H]
    \centering
    \includegraphics[width=0.75\linewidth]{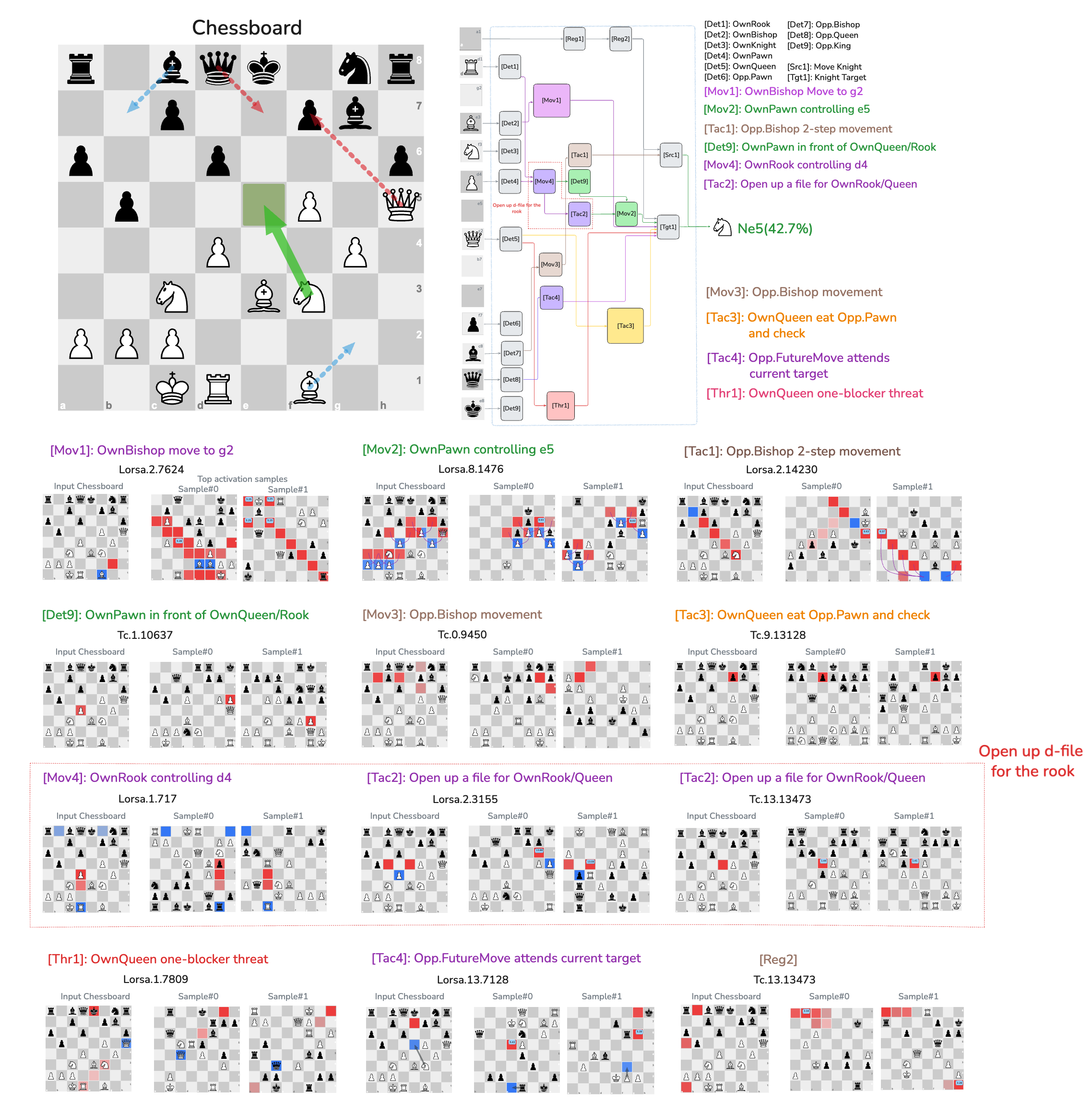}
    \caption{Reasoning pathway for a more complex position. BT4 correctly selects \textcolor{BestMove}{\textbf{\texttt{Ne5}}} with probability 42.7\%. The pathway suggests that \textcolor{BestMove}{\textbf{\texttt{Ne5}}} opens the d-file for the rook on d1, creates mating pressure by enabling a future \textbf{\texttt{Qf7+}}, and supports the development of \textbf{\texttt{Bg2}}. After \textcolor{BestMove}{\textbf{\texttt{Ne5}}}, \textbf{\texttt{Bb7}} no longer attacks the knight along the diagonal, and Black may respond with \textbf{\texttt{Qe7}}.}
    \label{fig:grandmaster-level_sacrificing}
\end{figure}

In the case shown in Figure~\ref{fig:grandmaster-level_sacrificing}, \textcolor{BestMove}{\textbf{\texttt{Ne5}}} is the only move that preserves White's advantage. Even strong human players, such as our 2100+ Elo annotator, may fail to find this sacrifice and instead prefer a more natural developing move such as \textbf{\texttt{Bg2}}. However, BT4 correctly assigns the highest probability to \textcolor{BestMove}{\textbf{\texttt{Ne5}}}. The reasoning pathway suggests that the model's mechanistic considerations behind choosing this move include the following: \textcolor{BestMove}{\textbf{\texttt{Ne5}}} helps open the d-file for the rook on d1, creates mating pressure by enabling a future \textbf{\texttt{Qf7+}}, and supports the subsequent development of \textbf{\texttt{Bg2}}. At the same time, after \textcolor{BestMove}{\textbf{\texttt{Ne5}}}, \textbf{\texttt{Bb7}} no longer attacks the knight along the diagonal, and the model also takes into account plausible opponent replies such as \textbf{\texttt{Qe7}}. Overall, this case illustrates how the model combines multiple positional and tactical factors when evaluating a non-obvious sacrifice.

\end{document}